\documentclass[3p,12pt]{elsarticle}

\journal{International Journal of Approximate Reasoning}

\usepackage{amssymb}
\usepackage{amsmath}
\usepackage{subfigure}
\usepackage{bbm}
\usepackage{graphicx}
\usepackage{algorithmic}
\usepackage{algorithm}
\usepackage{environ}
\usepackage{commath}
\usepackage{rotating}
\usepackage{eurosym}
\usepackage{url}

\usepackage{color}
\usepackage{ulem}
\newcommand{\reels}{\mathbb{R}}

\newcommand{\esp}{\mathbb{E}}

\newcommand{\Einf}{\underline{\esp}}
\newcommand{\Esup}{\overline{\esp}}

\newcommand{\cinf}{\underline{c}}
\newcommand{\csup}{\overline{c}}

\newcommand{\calP}{{\mathcal P}}

\newcommand{\calL}{{\mathcal L}}

\newcommand{\calF}{{\mathcal F}}

\newcommand{\calC}{\mathcal{C}}

\newcommand{\calM}{\mathcal{M}}

\newcommand{\calH}{\mathcal{H}}

\def\pref{\succcurlyeq}

\def\bomega{\boldsymbol{\omega}}

\def\blambda{\boldsymbol{\lambda}}

\def\Xi{\boldsymbol{\xi}}

\def\w{{\boldsymbol{w}}}

\def\a{{\boldsymbol{a}}}
\def\p{{\boldsymbol{p}}}

\newcommand{\bi}{\begin{itemize}}
\newcommand{\ei}{\end{itemize}}
\newcommand{\be}{\begin{enumerate}}
\newcommand{\ee}{\end{enumerate}}
\newcommand{\bd}{\begin{description}}
\newcommand{\ed}{\end{description}}

\newtheorem{Exp}{Example}
\newtheorem{Theo}{Theorem}
\newtheorem{Rem}{Remark}

\newcommand{\geinf}{\geqslant}
\newcommand{\gesup}{\eqslantgtr}

\newcommand{\gedr}{\gg}
\newcommand{\gew}{\gtrsim}

\NewEnviron{eqs}
{
\begin{subequations}
\begin{align}
\BODY
\end{align}
\end{subequations}
}

\begin{document}

\begin{frontmatter}

\title{Decision-Making with Belief Functions: a Review\footnote{This paper has been published in {\em International Journal of Approximate Reasoning} 109:87--110, 2019.}}
 
\author{Thierry œDen{\oe}ux\corref{cor1}}
\ead{thierry.denoeux@utc.fr}

\address{Universit\'e de Technologie de Compi\`egne, CNRS \\
UMR 7253 Heudiasyc, Compi\`egne, France}

\begin{abstract}
Approaches to decision-making under uncertainty in the belief function framework are reviewed. Most methods are shown to blend  criteria for decision under ignorance with the maximum expected utility principle of Bayesian decision theory. A distinction is made between  methods that construct a complete preference relation among acts, and those that allow incomparability of some acts due to lack of information. Methods  developed in the imprecise probability framework are applicable in the Dempster-Shafer context and are also reviewed.  Shafer's constructive decision theory, which substitutes the notion of goal for that of utility, is described and contrasted with other approaches. The paper ends by pointing out the need to carry out deeper investigation of fundamental issues related to decision-making with belief functions and to assess the descriptive, normative and prescriptive values of the different approaches. 
\end{abstract}


\end{frontmatter}

\section{Introduction}

The idea of using completely monotone capacities, or \emph{belief functions}, to model uncertainty dates back to Dempster's seminal work on statistical inference \cite{dempster67a,dempster68a,dempster68b}. It was later elaborated by Shafer in his 1976 book \cite{shafer76,shafer16a}, to become a full-fledged theory of uncertainty, now commonly referred to as Dempster-Shafer (DS) theory, evidence theory, or theory of belief functions \cite{denoeux17b}. In short, DS theory starts with the definition of a \emph{frame of discernment} $\Omega$ containing all the possible values  some variable $X$ can take. One and only one element of $\Omega$ is the true value. Independent pieces by evidence about $X$ are then represented by  belief functions and combined using a suitable operator called \emph{Dempster's rule of combination}.  

Much of the appeal of this theory is due to the generality of the belief function framework. If $Bel$ is additive, it is a probability measure, and the usual probabilistic formalism if recovered. If there is some subset $A$ of $\Omega$ such that $Bel(B)=1$ if $B$ contains $A$ and $Bel(B)=0$ otherwise, then belief function $Bel$ represents a state of knowledge in which we know for sure that the truth is in $A$, and nothing else. In particular, the case $A=\Omega$ corresponds to complete ignorance or lack of evidence. Belief functions thus allow us to represent logical information, probabilistic information, or any combination of both. In that sense, belief functions can be seen both as generalized sets \cite{dubois86a}, and as generalized probability measures.

Whereas Shafer's book presented in great detail the mathematics of belief functions in the finite setting as well as  mechanisms for  combining  belief functions, possibly  expressed in different  frames, it remained silent on the important issue of decision-making.  Shafer wrote a paper on this topic in the early 1980's \cite{shafer16a}, but this paper remained unpublished until recently \cite{shafer16d}. In the last 40 years, many researchers have attempted to fill this vacuum and propose methods for making decisions when uncertainty is described by belief functions. The objective of this review paper is to provide a broad picture of these endeavors in view of clarifying the main issues and indicating directions for further research.  

As belief functions make it possible to represent both ignorance and probabilistic information, most approaches to decision-making using belief functions extend classical methods for making decision under ignorance or  probabilistic uncertainty. We will thus start with a  brief reminder of these classical methods in Section \ref{sec:classic}, after which the belief function framework will be recalled in Section \ref{sec:DS}. We will then proceed with a systematic exposition of decision methods in the belief function framework. Criteria for decision-making with belief functions directly extending the classical criteria will first be reviewed  in Section \ref{sec:expect}, and approaches based on the imprecise-probability  view of belief functions will be described  in Section \ref{sec:IP}.  Shafer's ``constructive''  decision theory  \cite{shafer16d}, in which the concept of ``goal'' replaces that of ``utility'' will be exposed in Section \ref{sec:goals}.  Finally, the different approaches will be summarized and discussed  in Section \ref{sec:concl}.

\section{Classical Decision Theories}
\label{sec:classic}

In this section, we will first introduce the formal setting as well as the main notations and definitions in Section \ref{subsec:defi}. The two classical frameworks of \emph{decision under ignorance} and \emph{decision under probabilistic uncertainty} will then be recalled, respectively,  in Sections \ref{subsec:ignorance} and \ref{subsec:proba}.

\subsection{Definitions and Notations}
\label{subsec:defi}

A decision problem can be seen as a situation in which a decision-maker (DM) has to choose a course of action (or \emph{act}) in some set $\calF=\{f_1,\ldots,f_n\}$. An act may have different \emph{consequences}, depending on the \emph{state of nature}. Denoting by $\Omega=\{\omega_1,\ldots,\omega_s\}$ the set of states of nature and by $\calC=\{c_1,\ldots,c_r\}$ the set of consequences (or \emph{outcomes}), an act can thus be formalized as a mapping $f$ from $\Omega$ to $\calC$. In this paper, the three sets $\Omega$, $\calC$ and $\calF$ are assumed to be finite.

It is often assumed that the desirability of the consequences can be modeled by a \emph{quantitative utility function} \index{utility}
 $u: \calC \rightarrow \reels$, which assigns a numerical value to each consequence. The higher this value, the more desirable is the consequence for the DM.  Utilities can be elicited directly, or then can sometimes be deduced from the observation of the DM's preferences under uncertainty \cite{von_neumann44,savage54}. If the acts are indexed by $i$ and the states of nature by $j$, we will denote by $c_{ij}=f_i(\omega_j)$ the consequence  of selecting act $f_i$ if state $\omega_j$ occurs, and by $u_{ij}=u(c_{ij})$ the corresponding utility. The  $n\times s$ matrix $U=(u_{ij})$ will be called a \emph{payoff} or \emph{utility matrix}. These notions will now be illustrated using the following example inspired from \cite{jansen18}.

\begin{Exp}
\label{ex:invest}
Assume that the DM wants to invest money in stocks of some company. The acts then correspond to the stocks of the different companies. We assume that the DM considers five different stocks in $\calF = \{f_1,\ldots, f_5\}$. The states of nature  correspond to different economic scenarios that might  occur and which would influence the payoffs of the stocks of the different companies. Suppose that the DM considers three scenarios collected in $\Omega = \{\omega_1, \omega_2, \omega_3\}$. The payoff matrix is shown in Table \ref{tab:payoffs}. 

\begin{table}
\caption{Payoff matrix for the investment example. \label{tab:payoffs}}
\begin{center}
\begin{tabular}{cccc}
\hline
$u_{ij}$ & $\omega_1$ & $\omega_2$ & $\omega_3$\\
\hline
$f_1$ & 37 & 25 &23  \\
$f_2$ & 49 & 70 &2 \\
$f_3$ & 4 & 96 &1\\
$f_4$ &22 & 76 &25\\
$f_5$ &35 & 20 &23\\
\hline
\end{tabular}
\end{center}
\end{table}%

\end{Exp}

If the true state of nature $\omega$ were known, then the desirability of an act $f$ could be deduced from that of its consequence $f(\omega)$. Typically, however, the state of nature is unknown. A decision problem is then described by (1) the payoff matrix $U$ and (2) some description of the uncertainty  about the state of nature. The outcome of the decision problem is typically a preference relation $\pref$ among acts. This relation  is interpreted as follows: given two acts $f$ and $f'$, $f\pref f'$ means that $f$ is found by the DM to be \emph{at least as desirable} as $f'$. We also define the strict preference relation as $f \succ f'$ iff $f \pref f'$ and $\neg(f' \pref f)$ (meaning that $f$ is strictly more desirable than $f'$) and an indifference relation $f \sim f'$  iff $f \pref f'$ and $f' \pref f$ (meaning that $f$ and $f'$ are equally desirable). 

The preference relation is generally assumed to be reflexive (for any $f$, $f\pref f$) and transitive (for any $f,f',f''$, if  $f\pref f'$ and  $f'\pref f''$, then  $f\pref f''$): it is then a \emph{preorder}. If, additionally, the relation is antisymmetric (for any $f,f'$, if $f\pref f'$ and $f'\pref f$, then $f=f'$), then it is an \emph{order}. This preference relation is \emph{complete} if, for any two acts $f$ and $f'$, $f\pref f'$ or $f' \pref f$. Otherwise, it is \emph{partial}. An act $f$ is a \emph{greatest element} of relation $\pref$ if it is at least as desirable as any other act, i.e, if, for any $f'\in \calF$, $f\pref f'$. A complete preorder always has at least one greatest element, and it has only one if it is a complete order. An act $f$ is a \emph{maximal (or non-dominated) element} of the strict preference relation if no other act is strictly preferred to $f$, i.e., if for any $f'\in \calF$, $\neg (f' \succ f)$. A greatest element is a maximal element, but the converse is not true in general.

Most decision methods provide a complete or partial preorder of the set $\calF$ of acts. We can then compute the set of greatest elements in the former case, and the set of maximal elements in the latter. Some methods do not give us a preference relation, but directly a \emph{choice set}, defined as subset $\calF^* \subseteq \calF$ composed of the ``most preferred'' acts. We can then reconstruct a partial preference relation such that all elements in $\calF^*$ are greatest elements as follows:
\[
\forall f,f' \in \calF^*, \quad f \sim f' 
\]
\[
\forall f\in \calF^*,  \forall f'\not\in \calF^*,\quad f \succ f'.
\]

\subsection{Decision under Ignorance}
\label{subsec:ignorance}

Let us start with the situation where the DM is totally ignorant of the state of nature. All the information given to the DM is thus the utility matrix $U$.  A act $f_i$ is said to be \emph{dominated} by $f_k$ if the consequences of act $f_k$ are always at least as desirable as those of act $f_i$, whatever the state of nature, and strictly more desirable in at least one state, i.e., if $u_{ij} \le u_{kj}$ for all $j$, and $u_{ij}< u_{kj}$ for some $j$. According to the \emph{non-domination principle }\cite{szaniawski60}, an act that is dominated by another one should never be chosen and can, therefore, be discarded. For instance, in Table \ref{tab:payoffs}, we can see that act $f_5$ is dominated by $f_1$: consequently, we can remove $f_5$ from further consideration. 

After all dominated acts have been removed, there remains the problem of ordering the non-dominated acts by desirability, and  finding the set of most desirable acts. In the following, we first recall some classical decision methods in this setting, as well as a more recent generalization. We then discuss some axiomatic arguments proposed by Arrow and Hurwicz \cite{arrow77}. 

\subsubsection*{Classical Criteria}

 Several criteria of ``rational choice'' that have been proposed to derive a complete preference relation over acts. They are summarized in the following list (see, e.g., \cite{luce57, szaniawski60}):

\bi
\item The \emph{maximax criterion} considers, for each act, its more favorable consequence. We then have $f_i \pref f_{k}$  iff 
\begin{equation}
 \max_j u_{ij} \ge  \max_j u_{kj}.
\end{equation}
\item Conversely, Wald's \emph{maximin criterion} \cite{wald45} takes into account the least favorable consequence of each act: act $f_i$ is thus at least as desirable as $f_k$ iff
\begin{equation}
\min_j u_{ij} \ge  \min_j u_{kj}.
\end{equation}
\item The \emph{Hurwicz criterion} \cite{hurwicz51} considers, for each act,  a convex combination of the minimum and maximum utility: $f_i \pref f_{k}$ iff 
\begin{equation}
\label{eq:hurwicz}
\alpha \min_j u_{ij} + (1-\alpha)\max_j u_{ij} \ge  \alpha \min_j u_{kj} + (1-\alpha)\max_j u_{kj},
\end{equation} 
where $\alpha$ is a parameter in $[0,1]$ called the \emph{pessimism index}.
\item The \emph{Laplace criterion} ranks acts according to the average utility of their consequences: $f_i \pref f_{k}$ iff 
\begin{equation}
\frac{1}{s} \sum_{j=1}^s u_{ij} \ge  \frac{1}{s} \sum_{j=1}^s u_{kj}.
\end{equation}
\item Finally, the \emph{minimax regret criterion} \cite{savage51} considers an act $f_i$ to be at least as desirable as $f_k$ if it has smaller maximal regret, where \emph{regret} is defined as the utility difference with the best act, for a given state of nature. More precisely, let the regret $r_{ij}$ for act $f_i$ and state $\omega_j$ be defined as follows,
\begin{equation}
\label{eq:regret}
r_{ij}=\max_\ell u_{\ell j}-u_{ij}.
\end{equation}
The maximum regret for act $f_i$ is $R_i=\max_j r_{ij}$, and act $f_i$ is considered to be at least as  desirable as $f_k$ when $R_i \le R_k$.
\ei

\begin{Exp}
\label{ex:invest1}
Consider the payoff matrix of  Example \ref{ex:invest}. We have seen that act $f_5$ is dominated and should be ruled out. From the calculations shown in Tables \ref{tab:crit_ignorance} and \ref{tab:maxregret}, we can see that the above five criteria yield different strict preference relations:
\bi
\item Maximin: $f_1\succ f_4 \succ f_2 \succ f_3$
\item Maximax: $f_3\succ f_4 \succ f_2 \succ f_1$
\item Hurwicz with $\alpha=0.5$: $f_4\succ f_3 \succ f_2 \succ f_1$
\item Laplace: $f_4\succ f_2 \succ f_3 \succ f_1$
\item Minimax regret: $f_2\succ f_4 \succ f_3 \succ f_1$
\ei  

\begin{table}
\begin{center}
\caption{Calculation of the preference relations for the maximin, maximax, Hurwicz ($\alpha=0.5$) and Laplace criteria with the payoff matrix of Example \ref{ex:invest}. \label{tab:crit_ignorance}}
\begin{tabular}{cccccccc}
\hline
 & $u_{i1}$ & $u_{i2}$ & $u_{i3}$ & $\min_j u_{ij}$ & $\max_j u_{ij}$ & $0.5(\min_j u_{ij} + \max_j u_{ij})$ & $\frac{1}{s} \sum_j u_{ij}$\\
\hline
$f_1$ & 37 & 25 &23 & \textbf{23}&37&30&28.3  \\
$f_2$ & 49 & 70 &2&2&70&36&40.3 \\
$f_3$ & 4 & 96 &1&1& \textbf{96}&48.5&33.7\\
$f_4$ &22 & 76 &25&22&76&\textbf{49}&\textbf{41}\\
\hline
\end{tabular}
\end{center}
\end{table}

\begin{table}
\begin{center}
\caption{Calculation of the preference relation for the max regret criterion with the payoff matrix of Example \ref{ex:invest}. \label{tab:maxregret}}

\begin{tabular}{ccccccccc}
\hline
 & $u_{i1}$ & $u_{i2}$ & $u_{i3}$ &  $r_{i1}$ & $r_{i2}$&  $r_{i3}$ & $\max_j r_{ij}$\\
\hline
$f_1$ & 37 & 25 &23 & 12& 71 & 2& 71\\
$f_2$ & 49 & 70 &2 & 0& 26& 23 & \textbf{26}\\
$f_3$ & 4 & 96 &1& 45 & 0& 24 & 45\\
$f_4$ &22 & 76 &25& 27& 20& 0 & 27\\
\hline
\end{tabular}
\end{center}
\end{table}

\end{Exp}

The maximax and maximin criteria correspond, respectively, to extreme optimistic and pessimistic (or conservative) attitudes of the DM.  The Hurwicz criterion allows us to parameterize the DM's attitude toward ambiguity, using the pessimism index. Figure \ref{fig:hurwicz} shows the aggregated utilities as functions of the pessimism index. The Laplace criterion can be seen as an application of the expected utility principle (see Section \ref{subsec:proba} below), using a uniform probability distribution over the state of nature as an application of Laplace's principle of indifference. These four criteria amount to extending the utility function to sets, i.e., they aggregate, for each act $f_i$, the utilities $u_{ij}$ for all $j$, into a single number. The minimax regret criterion works differently, as it measures the desirability of an act by a quantity that depends on the consequences of all  other acts.

\subsubsection*{Ordered Weighted Average Criterion}

The Laplace, maximax, maximin and Hurwicz criteria correspond to different ways of aggregating  utilities  using, respectively, the average, the maximum, the minimum, and a convex sum of the minimum and the maximum. These four operators happen to belong to the family  of so-called \emph{Ordered Weighted Average (OWA)} operators \cite{yager88b}. An OWA operator of arity $s$ is a function $F:\reels^s \rightarrow \reels$ of the form
\begin{equation}
\label{eq:owa}
F(x_1,\ldots,x_s)=\sum_{i=1}^s w_i x_{(i)},
\end{equation}
where $x_{(i)}$ is the $i$-th largest element in the collection $x_1,\ldots,x_s$, and $w_1,\ldots,w_s$ are positive weights that sum to 1. It is clear that the four above-mentioned operators are obtained for different choices of the weights:
\bd
\item[Average:] $(1/s,1/s,\ldots,1/s)$;
\item[Maximum:] $(1,0,\ldots,0)$;
\item[Minimum:] $(0,\ldots,0,1)$;
\item[Hurwicz:] $(1-\alpha,0,\ldots,0,\alpha)$.
\ed
In a decision-making context, each weight $w_i$ may be interpreted as a probability that the $i$-th best outcome will happen. Yager \cite{yager88b} defines the \emph{degree of optimism} of an OWA operator with weight vector $\w$ as
\begin{equation}
\label{eq:optimism}
OPT(\w)=\sum_{i=1}^s  \frac{s-i}{s-1} w_i.
\end{equation}
The degree of optimism equals 1  for the maximum,  0 for the minimum, 0.5 for the mean, and $1 - \alpha$ for the Hurwicz criterion. Given a degree of optimism $\beta$, Yager \cite{yager88b} proposes to choose the OWA operator $F_\beta$ that maximizes the entropy
\begin{equation}
ENT(\w)=-\sum_{i=1}^s w_i \log w_i,
\end{equation}
under the constraint $OPT(\w)=\beta$.

\begin{Exp}
\label{ex:OWA}
Consider again the data of Example \ref{ex:invest}. With $\beta=0.2$ and $\beta=0.7$, we get, respectively, $w=(0.0819, 0.236, 0.682)$ and $w=(0.554, 0.292, 0.154)$. The aggregating utilities for these two cases are shown in Table \ref{tab:OWA}, and  the corresponding preference relations are:
\bi
\item $\beta=0.2$: $f_4 \succ f_1 \succ f_2 \succ f_3$
\item $\beta=0.7$: $f_3 \succ f_2 \succ f_4 \succ f_1$.
\ei
Figure \ref{fig:OWA} shows the aggregated utilities for each of the four acts, as functions of $\beta$. Comparing Figures \ref{fig:hurwicz} and \ref{fig:OWA}, we can see that the Hurwicz and OWA criteria yield similar results in this case. However, the OWA parametrization allows us to recover the Laplace criterion for $\beta=0.5$.
      
\begin{table}
\begin{center}
\caption{Aggregated utilities using the OWA aggregation operator with $\beta=0.2$ and $\beta=0.7$ and  the payoff matrix of Example \ref{ex:invest}. \label{tab:OWA}}
\begin{tabular}{cccccc}
\hline
 & $u_{i1}$ & $u_{i2}$ & $u_{i3}$ & $F_{0.2}(u_{i1},u_{i2},u_{i3})$ & $F_{0.7}(u_{i1},u_{i2},u_{i3})$\\
\hline
$f_1$ & 37 & 25 &23 & 24.62 &31.34 \\
$f_2$ & 49 & 70 &2&18.67 & 53.40\\
$f_3$ & 4 & 96 &1&9.49& \textbf{54.50}\\
$f_4$ &22 & 76 &25&\textbf{27.13}& 52.79\\
\hline
\end{tabular}
\end{center}
\end{table}

\begin{figure}
\centering 
\centering 
\subfigure[\label{fig:hurwicz}]{\includegraphics[width=0.49\textwidth]{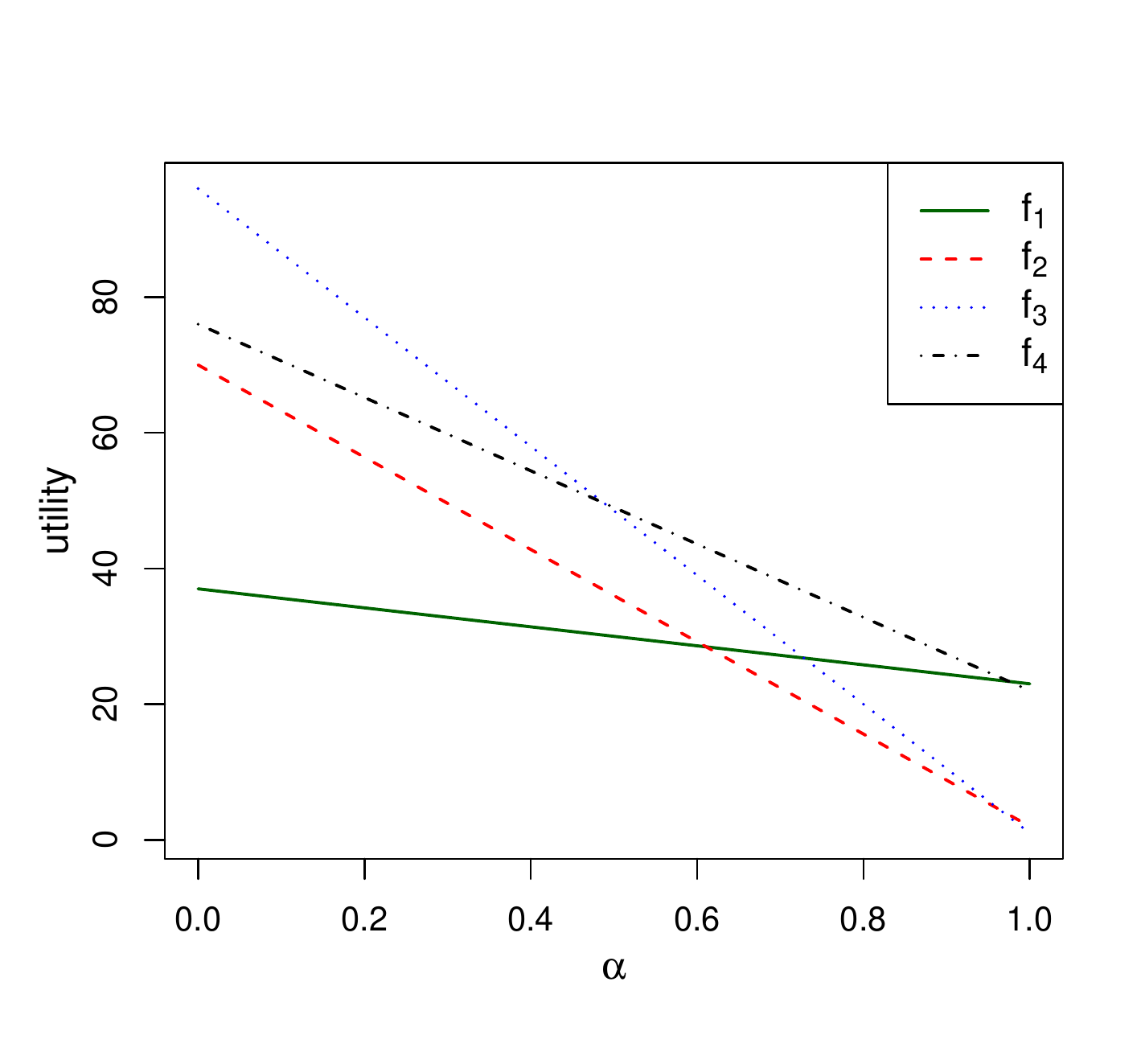}}
\subfigure[\label{fig:OWA}]{\includegraphics[width=0.49\textwidth]{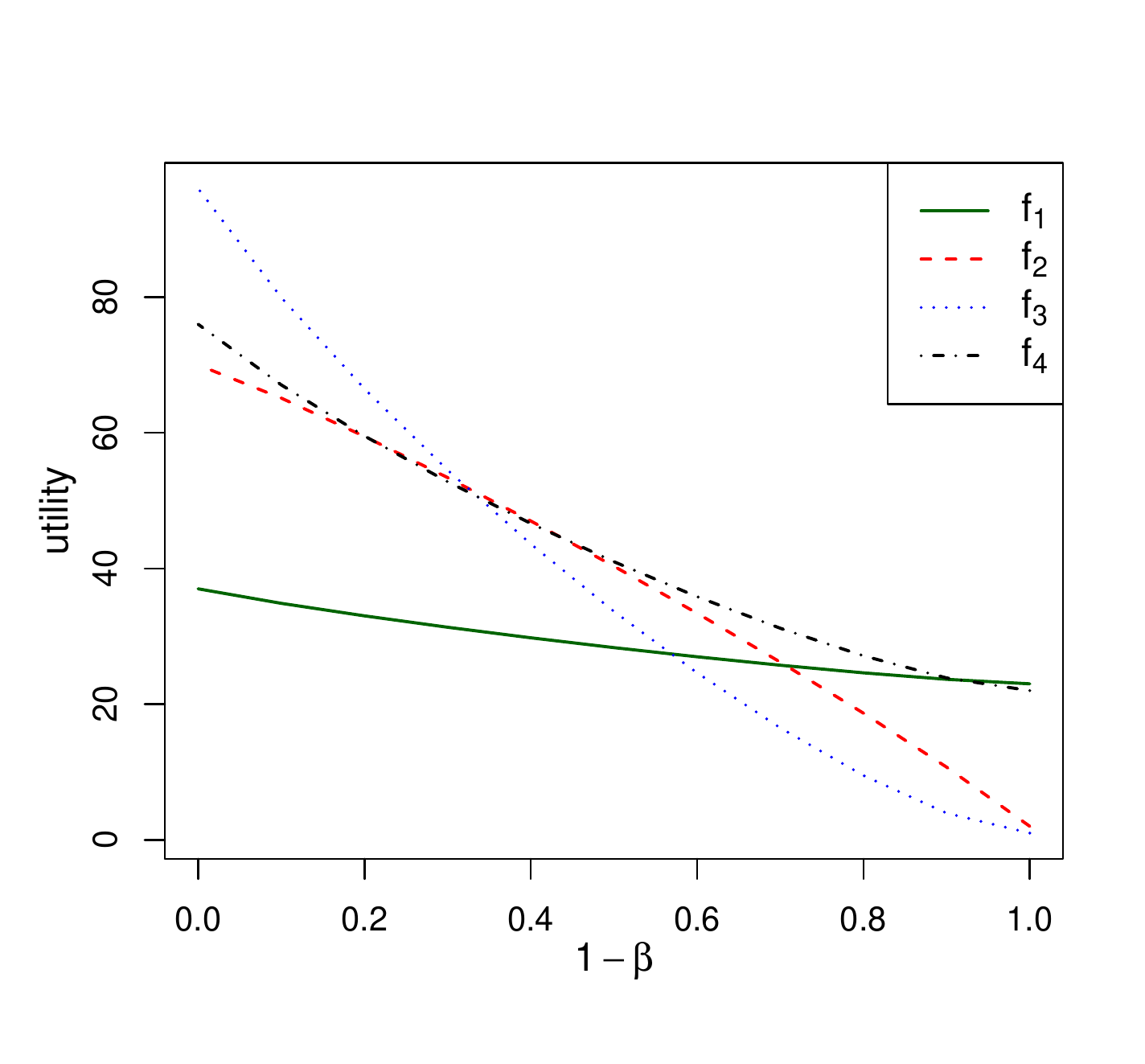}}
\caption{Aggregated utilities vs. pessimism index $\alpha$ for the Hurwicz criterion (a) and vs. one minus the degree of optimism $\beta$ for the OWA criterion (b).  (Example  \ref{ex:OWA}). \label{fig:hurwicz_OWA}}
\end{figure}

\end{Exp}
\subsubsection*{Axiomatic Arguments}

The fact that different criteria yield different (and, sometimes, even opposite) results is disturbing and it calls for axiomatic arguments to support the choice of a criterion. Given a set of acts $\calF$, each of the above five criteria  induces a complete preorder $\pref$  and a choice set $\calF^*$ containing the greatest elements of $\pref$. Arrow and Hurwicz \cite{arrow77} have proposed four axioms that a choice operator $\calF \rightarrow \calF^*$ (i.e., a way of constructing the choice set) should verify. The following description of these axioms is taken from \cite{giang12}.
\bd
\item[Axiom $A_1$:] The non-empty intersection of a decision problem (set of acts) and the choice set of a larger decision problem is the choice set of the former. Formally, if $\calF_1 \subset \calF_2$ and $\calF_2^* \cap \calF_1\neq \emptyset$, then $\calF_1^*=\calF_2^* \cap \calF_1$. 
\item[Axiom $A_2$:] Relabeling actions and states does not change the optimal status of actions. Formally, if $\phi_a$ is a one-to-one mapping from $\calF_1$ to $\calF_2$ and  $\phi_s$ is a one-to-one mapping from $\Omega_1$ to $\Omega_2$ such that, for all $f\in \calF_1$ and for all $\omega\in \Omega_1$, $f(\omega)=\phi_a(f)(\phi_s(\omega))$, then $f \in \calF^*_1$ iff $\phi_a(f) \in \calF^*_2$.
\ed
Given a set of acts $\calF$, a state $\omega\in \Omega$ is said to be \emph{duplicate} if there exists another state $\omega'$ in $\Omega$ such that, for all $f\in \calF$, $f(\omega)= f(\omega')$. Deleting a duplicate state $\omega$ means defining a new state space $\Omega'=\Omega \setminus \{\omega\}$ and the new set of actions $\calF|_{\Omega'}$ containing the restrictions $f|_{\Omega'}$ of all acts $f$ in $\calF$. We then have the following postulate.

\bd
\item[Axiom $A_3$:] Deletion of a duplicate state does not change the optimality status of actions. Formally, $f\in \calF^*$ iff $f|_{\Omega'} \in (\calF|_{\Omega'})^*$.
\item[Axiom $A_4$ (dominance):] If $f\in \calF^*$ and $f'$ dominates $f$, then $f'\in \calF^*$. If $f\not \in \calF^*$ and $f$ dominates $f'$, then $f' \not\in \calF^*$.
\ed

Axiom $A_1$ is clearly violated by the minimax regret criterion. To see this, consider again, for instance, the decision problem of Example \ref{ex:invest}. For $\calF_1=\{f_1,f_2,f_3,f_4\}$, we have seen that $\calF_1^*=\{f_2\}$ according to the minimax regret criterion. Now, consider a new act $f_6$ such that $u_{61}=0$, $u_{62}= 100$ and $u_{63}=0$, and $\calF_2=\{f_1,f_2,f_3,f_4,f_6\}$. As a consequence of the introduction of  this new act, the maximal regrets of $f_1$, $f_2$, $f_3$, $f_4$ and $f_6$ now  become, respectively,  75, 30, 45, 27  and 49. Hence, $\calF_2^*=\{f_4\}$ and $\calF_1^*\neq \calF_2^*\cap \calF_1$. Act $f_4$, which was initially considered strictly less desirable than $f_2$, becomes strictly more desirable after an additional act $f_6$ is considered.

It is also easy to see that Axiom $A_3$ is violated by the Laplace criterion. To illustrate this point, assume that, in Example  \ref{ex:invest}, we split the state of nature $\omega_1$ in two states: ``Economic scenario 1 occurs and there is life on Mars'' ($\omega'_1$) and ``Economic scenario 1 occurs and there is no life on Mars'' ($\omega''_1$). It is clear that the payoffs of the real estate investments are identical under $\omega'_1$ and $\omega''_1$. Consequently, the new payoff matrix will be obtained by duplicating the first column in Table \ref{tab:payoffs}. With this  payoff matrix, the average utilities for the four acts are 30.5, 42.5, 26.25 and 36.25. Consequently, the choice set for the Laplace criterion is $\calF^*= \{f_2\}$. Should we learn that there is no life of Mars, we would delete $\omega'_1$ and get that same payoff matrix as in Table \ref{tab:payoffs}, resulting in choice set  $\{f_4\}$. Learning that there is no life on Mars thus made us  change our investment decision!  

Consequently, convincing  arguments can be put forward for rejecting  the La\-place and minimax regret criteria  as criteria for rational decision-making under ignorance. A stronger result, due to Arrow and Hurwicz \cite{arrow77} is that, under some regularity assumptions, Axioms $A_1$ to $A_4$ imply that the choice set depends only on the worst and the best consequences of each act. This result provides a strong argument in favor of  the Hurwicz criterion (\ref{eq:hurwicz}).

\subsection{Decision under Probabilistic Uncertainty}
\label{subsec:proba}

Let us now consider the situation where uncertainty about the state of nature is quantified by probabilities $p_1,\ldots,p_s$ on $\Omega$. Typically, these probabilities are assumed to be objective: we say that we have a problem of \emph{decision under risk}. However, the following developments also apply to the case where the probabilities are subjective. In any case, the probability distribution $p_1,\ldots,p_s$ is assumed to be known, together with the utility matrix $U$. We can then compute, for each act $f_i$, its  expected utility as
\begin{equation}
EU(f_i)=\sum_{j=1}^s u_{ij} p_j.
\end{equation}
According to the Maximum Expected Utility (MEU) principle, an act $f_i$ is more desirable than an act $f_k$ if its yields more desirable consequences \emph{on average} over all possible states of nature, i.e.,  if it has a higher expected  utility: $f_i \pref f_k$ iff $EU(f_i) \ge EU(f_k)$. 

\subsubsection*{Axiomatic justification}

The MEU principle was first axiomatized by von Neumann and Morgenstern \cite{von_neumann44}. We give hereafter a summary of their argument. Given a probability distribution on $\Omega$, an act $f:\Omega \rightarrow \calC$ induces a probability measure $P$ on the set $\calC$ of consequences (assumed to be finite), called a \emph{lottery}. We denote by $\calL$  the set of lotteries on $\calC$. 
If we agree that two acts providing the same lottery are equivalent, then the problem of comparing the desirability of acts becomes that of comparing the desirability of lotteries. Let $\pref$ be   a preference relation among lotteries. Von Neumann and Morgenstern argued that, to be rational, a preference relation should verify the following three axioms.

\bd
\item[Complete preorder:] the preference relation is a complete and non-trivial preorder (i.e., it is a reflexive, transitive and complete relation) on $\calL$.
\item[Continuity:] for any lotteries $P$, $Q$ and $R$ such that $P \succ Q \succ R$, there exist  probabilities $\alpha$ and $\beta$ in $(0,1)$ such that
\begin{equation}
\alpha P + (1-\alpha) R \succ Q \succ \beta P + (1-\beta) R,
\end{equation}
where $\alpha P + (1-\alpha) R$ is a compound lottery, which refers to the situation where you receive $P$ with probability $\alpha$ and $Q$ with probability $1-\alpha$. This axiom means that (1) no lottery $R$  is so undesirable that it cannot become desirable if mixed with some very desirable lottery $P$, and (2)  that, conversely,  no act $P$ is so desirable that it cannot become undesirable if mixed with some very undesirable lottery $R$.
\item[Independence:] for any lotteries $P$, $Q$ and $R$ and for any $\alpha\in (0,1]$,
\begin{equation}
P \pref Q \Leftrightarrow \alpha P + (1-\alpha) R  \pref \alpha Q + (1-\alpha) R.
\end{equation}
\ed

We then have the following theorem.

\begin{Theo}[Von Neumann and Morgentern] The two following propositions are equivalent:
\be
\item The preference relation $\succeq$ verifies the axioms of complete preorder, continuity, and independence;
\item There exists a utility function $u:\calC \rightarrow \reels$ such that, for any two lotteries $P=(p_1,\ldots,p_r)$ and $Q=(q_1,\ldots,q_r)$,
\begin{equation}
P \succeq Q \Leftrightarrow \sum_{i=1}^r p_i u(c_i) \ge \sum_{i=1}^r q_i u(c_i).
\end{equation}
\ee
Function $u$ is unique up to a strictly increasing affine transformation. 
\end{Theo}

\subsubsection*{Discussion}

Von Neumann and Morgenstern's theorem has had a tremendous impact, as it provides a compelling justification of both the notion of utility, and the MEU principle. For problems of decision under risk, the normative value of the MEU principle is widely accepted. From a descriptive point of view, violations of the MEU principle by most DMs in some particular situations have been demonstrated  experimentally by Allais \cite{allais53}, among others. 

For problems of decision under uncertainty (in which probabilities are not given in advance), Savage \cite{savage54} has argued, based on rationality requirements, that a  DM should \emph{always} maximize expected utility, for some subjective probability measure and utility function. However, the relevance of Savage's axioms has been questioned (see, e.g., \cite{shafer86}). Moreover, Ellsberg \cite{ellsberg61} has shown experimentally that, in the presence of ambiguity,  people tend to make decisions  in a way that  is  not  consistent with the ``sure thing principle'', one of Savage's axioms. Ellsberg's paradox has sparkled a rich literature in theoretical economics aiming to derive axioms that result in decision rules that better describe the way humans make decision when ambiguity is present. For instance, Gilboa \cite{gilboa87} and Schmeidler \cite{schmeidler89} have derived axioms that justify making decisions by maximizing the Choquet expectation of a utility function with respect to a non-additive measure.  As we will see in Section \ref{subsec:upperlower}, these results are consistent with some decision rules that have been proposed in the DS framework.  

\subsubsection*{Statistical preference and stochastic dominance}

Given an act $f$, a probability measure $P$ on $\Omega$ and a utility function, we can define the real random variable $X=u\circ f$, which is called a \emph{gamble} \cite{walley91}. Gamble $X$ represents the uncertain utility we get if we select act $f$.  Let us now consider two gambles $X$ and $Y$ induced by two acts defined on the same state space $\Omega$. According to the MEU, $X$ is at least as desirable as $Y$ iff $\esp(X) \ge \esp(Y)$.  As noted in by Couso and Dubois \cite{couso12} and Couso \cite{couso14}, this ``dominance in expectation'' relation is just one way to compare two random variables. Two well-known alternatives are the \emph{statistical preference} and \emph{first-order stochastic dominance} relations.

We say that  $X$ is \emph{statistically preferred} to $Y$ (and we note $X \succeq_{SP} Y$) if $P(X>Y)\ge P(X<Y)$ or, equivalently, $P(X>Y)+\frac{1}{2} P(X=Y) \ge 0.5$. We note $X \succ_{SP} Y$ if $X \succeq_{SP} Y$ but $\neg (Y \succeq_{SP} X)$. The statistical preference relation is complete, but it is not transitive: it is possible to have three gambles $X$, $Y$ and $Z$ such that  $X \succ_{SP} Y$ and  $Y \succ_{SP} Z$, but  $Z \succ_{SP} X$ \cite{deschuymer03}. This lack of transitivity makes the usefulness of the statistical preference relation questionable for decision-making.

The notion of (first-order) stochastic dominance \cite{whitt88} seems to be more useful. We say that $X$ is \emph{stochastically greater} than $Y$, and we write $X \succeq_{SD} Y$ if, for any $x\in \reels$, 
\begin{equation}
\label{eq:StochD}
P(X>x) \ge P(Y>x).
\end{equation}
 The meaning of this relation is clear: gamble $X$ is at least as desirable as gamble $Y$ if any utility threshold $x$ has a greater probability of being exceeded by $X$ than it has by $Y$. The stochastic relevance relation is a partial order. Also, it is well-known that $X \succeq_{SD} Y$ iff $\esp[h(X)] \ge \esp[h(Y)]$ for any bounded non decreasing function $h:\reels \to \reels$ \cite{lehman55}. Consequently,  stochastic relevance is particularly relevant in situations where the utility function is only known up to a non decreasing transformation.

Extensions of the statistical preference and stochastic dominance relations in the DS setting will be discussed in Section \ref{subsec:partial}.

\section{Theory of Belief Functions}
\label{sec:DS}

Before reviewing decision methods in the belief function framework in the following sections, we will first recall the main definitions and results pertaining to belief functions in Section \ref{subsec:rec_DS}. The motivation for considering decision problems in this framework will then be exposed in Section \ref{subsec:motiv}.

\subsection{Belief Functions}
\label{subsec:rec_DS}

\paragraph{Basic definitions} As before, let $\Omega$ be the set of states of nature. A \emph{mass function} \cite{shafer76} is a mapping $m$ from the power set of $\Omega$, denoted by $2^\Omega$, to $[0,1]$, such that
\[
\sum_{A \subseteq \Omega} m(A)=1,
\]
and $m(\emptyset)=0$. Any subset $A$ of $\Omega$ such that $m(A)>0$ is called a \emph{focal set} of $m$. In DS theory, $m$ is used as a representation of a piece of evidence about some variable $X$ taking values in $\Omega$. Such a function arises when we compare the evidence to the situation in which we receive a coded message \cite{shafer81}, and we know that the code has been selected at random from a set $S=\{s_1,\ldots,s_q\}$ with known probabilities $p_1,\ldots,p_q$. If code $s_i$ was selected, then the meaning of the message is $X\in \Gamma(s_i)$, where $\Gamma$ is a multi-valued mapping from $S$ to $2^\Omega$. In this setting,
\[
m(A)=\sum_{\{i \mid \Gamma(s_i)=A\}} p_i
\]
is the probability that the meaning of the code is $X \in A$, i.e., the probability of knowing only that $X\in A$, and nothing more. A mass function $m$ on a finite set $\Omega$ can always be seen as being induced by a probability space $(S,2^S,P)$, where $S$ is a finite set and $P$ a probability measure on $(S,2^S)$,  and a multi-valued mapping $\Gamma: S \to 2^\Omega$ \cite{dempster67a}.

Given a mass function $m$, \emph{belief} and \emph{plausibility} functions can be defined, respectively, as
\[
Bel(A)=\sum_{B \subseteq A} m(B)
\]
and 
\[
Pl(A)=\sum_{B \cap A\neq \emptyset} m(B)=1-Bel(\overline{A}),
\]
where $\overline{A}$ denotes the complement of $A$. The quantities $Bel(A)$ and $Pl(A)$ denote, respectively, the probability that the evidence implies the proposition $X\in A$, and the probability that the evidence does not contradict this proposition. Mathematically, a belief function is a completely monotone capacity, i.e., it verifies $Bel(\emptyset)=0$, $Bel(\Omega)=1$ and, for any $k\ge 2$ and for any family $A_1,\ldots,A_k$ of subsets of  $\Omega$,
\[
Bel\left( \bigcup_{i=1}^k A_i\right) \ge \sum_{\emptyset \neq I \subseteq \{1,\ldots,k\}} (-1)^{|I|+1} Bel\left( \bigcap_{i\in I} A_i \right).
\]  
Conversely, any completely monotone capacity $Bel$ corresponds a unique mass function $m$ such that
\[
 m(A)=\sum_{\emptyset \neq B \subseteq A} (-1)^{|A|-|B|} Bel(B) ,
\]
for all $A\subseteq \Omega$. 

\paragraph{Relationship with probabilistic and set-theoretic formalisms} When the focal sets of $m$ are singletons, functions $Bel$ and $Pl$ boil down to a single probability measure. Mass function $m$ is then said to be \emph{Bayesian}. DS theory is, thus, strictly more expressive than probability theory, which is recovered as a special case when the available information is uncertain, but precise. When there is only one focal set $A$, then $Bel(B)=I(A\subseteq B)$, where $I(\cdot)$ is the indicator function. Such a belief function is said to be \emph{logical}. It describes to a piece of evidence  that tell us that $X \in A$ for sure, and nothing more: it thus describes certain, but imprecise information.  There is a one-to-one correspondence between subsets and $\Omega$ and logical belief functions.  A general belief function (or, equivalently, its associated mass function) can, thus, be seen as a \emph{generalized set} \cite{dubois86a}.

\paragraph{Imprecise-probability view} Given a belief function $Bel$ induced by a mass function $m$, we can consider the set $\calP(m)$ of probability measures $P$ that dominate it, i.e., such that $P(A) \ge Bel(A)$ for all $A\subseteq \Omega$. Any such probability measure is said to be \emph{compatible} with $Bel$, and  $\calP(m)$ is called the \emph{credal set} of $m$. It is clear that this set is convex. An arbitrary element of $\calP(m)$ can be obtained by distributing each mass $m(A)$ among the elements of $A$. More precisely, let us call an \emph{allocation}  of $m$ any function 
\begin{equation}
\label{eq:alloc}
a: \Omega \times (2^\Omega\setminus\{\emptyset\}) \rightarrow[0,1] 
\end{equation}
such that, for all $A\subseteq \Omega$,
\begin{equation}
\sum_{\omega\in A} a(\omega,A)=m(A).
\end{equation}
Each quantity $a(\omega,A)$ can be viewed as a part of $m(A)$ allocated to the element $\omega$ of $A$. By summing up the numbers $a(\omega,A)$ for each $\omega$, we get a probability mass function on $\Omega$, 
\begin{equation}
p_a(\omega)=\sum_{A \ni \omega} a(\omega,A).
\end{equation}
It can be shown \cite{dempster67a} that the set of probability measures constructed in that way is exactly equal to the credal set $\calP(m)$. Furthermore, the following equalities hold for any $A\subseteq \Omega$:
\begin{align*}
Bel(A)&=\min_{P \in\calP(m)} P(A)\\
Pl(A)&=\max_{P \in\calP(m)} P(A).
\end{align*}
A belief function is, thus, a \emph{coherent lower probability}. However, not all coherent lower probabilities are belief functions \cite{walley91}. It must be emphasized that DS theory and the theory of imprecise probabilities (IP)  initiated by Walley \cite{walley91} and developed by his followers (see e.g., \cite{augustin14}) are different theoretical frameworks. In particular, the two theories have different rules of conditioning \cite{halpern92,jaffray92}. In Section \ref{sec:IP}, we will review some decision rules that have been proposed in the IP setting, because some of these rules can also be interpreted from the DS perspective, while others may receive such an interpretation in the future.

\subsection{Necessity  of a Theory of Decision-Making with Belief Functions}
\label{subsec:motiv}

As shown in the previous section, the formalism of belief functions, having more degrees of freedom than probability theory, allows for the representation of weaker forms of information, up to total ignorance. Belief functions  thus appear in decision problems when information is weaker than generally assumed in the probabilistic framework. In particular, two non exclusive situations are typically encountered.

The first situation is one in which the DM's information concerning the possible states of nature is best described by a mass function $m$ on $\Omega$. This is the case, for instance, in classification problem, when a classifier quantifies the uncertainty about the class of an object by a belief function, and a decision regarding the assignment of that object has to be made \cite{denoeux96b}. Any act $f$ then carries $m$ to the set $\calC$ of consequences. The mass  assigned  to each focal set $A$ of $m$ is transferred to  the image of $A$ by $f$, denoted as $f[A]=\{c\in \calC \mid f(\omega)=c \text{ for some } \omega\in A\}$. The resulting mass function\footnote{In the rest of this paper, we will denote mass functions on $\Omega$ by the letter $m$ and mass functions on $\calC$ (evidential lotteries) by the Greek letter $\mu$.} $\mu_f$ on $\calC$, called an \emph{evidential lottery},  is then defined by
\begin{equation}
\label{eq:decm1}
\mu_f(B) = \sum_{\{A \subseteq \Omega \mid f[A]=B\}} m(A),
\end{equation}
for any $B  \subseteq \calC$. 

The second situation in which belief functions come into the picture is that in which the consequences of each act under each state of nature may not be precisely described. As discussed in \cite{ghirardato01}, this situation may arise when the decision problem is underspecified: for instance, the set of acts $\calF$ or the state space $\Omega$ may be too coarsely defined.  In that case, an act may formally be represented by  a multi-valued mapping $f : \Omega \rightarrow 2^\calC$, assigning a \emph{set} of possible consequences $f(\omega) \subseteq \calC$ to each state of nature $\omega$. Given a probability distribution $p:\Omega \rightarrow [0,1]$,  $f$ then induces the following mass function $\mu_f$ on $\calC$,
\begin{equation}
\label{eq:decm2}
\mu_f(B) = \sum_{\{\omega \in \Omega \mid f(\omega)=B\}} p(\omega),
\end{equation}
for all $B \subseteq \calC$.

It is clear that these two situations can occur simultaneously, i.e., we may have a mass function $m$ on $\Omega$, and ill-known consequences. In that case, Equations  (\ref{eq:decm1}) and (\ref{eq:decm2}) become
\begin{equation}
\label{eq:decm3}
\mu_f(B) = \sum_{\{A \subseteq \Omega \mid \bigcup_{\omega\in A} f(\omega)=B\}} m(A),
\end{equation}
for all $B \subseteq \calC$. Assuming the (possibly Bayesian) mass function $m$ on $\Omega$ to be induced by a multi-valued mapping $\Gamma: S\to 2^\Omega$, where $S$ is a probability space,  mass function $\mu_f$ is induced by the multi-valued mapping $f^*\circ\Gamma: S \to 2^\calC$, where function $f^*:2^\Omega \to 2^\calC$ is defined as $f^*(A)=f[A]$ for any $A\subseteq \Omega$ if $f$ is single-valued and $f^*(A)=\bigcup_{\omega\in A} f(\omega)$ if $f$ is multi-valued.

\begin{Exp}
Let $\Omega=\{\omega_1,\omega_2,\omega_3\}$ and $m$ the following mass function on $\Omega$:
\begin{equation}
\begin{array}{ll}
m(\{\omega_1,\omega_2\})=0.3, & m(\{\omega_2,\omega_3\})=0.2, \\ 
m(\{\omega_3\})=0.4, & m(\Omega)=0.1.
\end{array}
\end{equation}
Let $\calC=\{c_1,c_2,c_3\}$ and $f$ the act
\begin{equation}
f(\omega_1)=\{c_1\}, \quad f(\omega_2)=\{c_1,c_2\}, \quad f(\omega_3)=\{c_2,c_3\}.
\end{equation}
To compute the induced mass function on $\calC$, we transfer the masses as follows:
\begin{eqs}
m(\{\omega_1,\omega_2\})=0.3 & \rightarrow f(\omega_1)\cup f(\omega_2)=\{c_1,c_2\}\\
m(\{\omega_2,\omega_3\})=0.2 & \rightarrow f(\omega_2)\cup f(\omega_3)=\{c_1,c_2,c_3\}\\
m(\{\omega_3\})=0.4 & \rightarrow  f(\omega_3)=\{c_2,c_3\}\\
m(\Omega)=0.1& \rightarrow f(\omega_1)\cup f(\omega_2)\cup f(\omega_3)=\{c_1,c_2,c_3\}.
\end{eqs} 
Finally, we obtain the following mass function on $\calC$:
\begin{equation}
\mu_f(\{c_1,c_2\})=0.3, \quad \mu_f(\{c_2,c_3\})=0.4, \quad \mu_f(\{c_1,c_2,c_3\})=0.3. 
\end{equation}
\end{Exp}

In any of the situations considered above, we can assign to each act $f$ an evidential lottery  $\mu_f$ on $\calC$. Determining preferences among acts then amounts to determining preferences among evidential lotteries. Several methods generalizing the decision criteria reviewed in Section \ref{sec:classic} will first be reviewed in Section \ref{sec:expect}. Decision methods based on the imprecise-probability view of belief functions will then be presented in Section \ref{sec:IP}.

\section{Extensions of Classical Criteria}
\label{sec:expect}

As recalled in Section \ref{subsec:rec_DS}, belief functions can be seen both as  generalized sets and as generalized probabilities. As a consequence, criteria for decision-making with belief functions can be constructed by blending the criteria for decision under ignorance reviewed in Section \ref{subsec:ignorance} with the MEU principle recalled in Section \ref{subsec:proba}. These criteria will be examined in Sections \ref{subsec:upperlower} to \ref{subsec:gen_minimax_regret}, and axiomatic arguments will be discussed in Section \ref{subsec:axiom}. Finally, partial preference relations among evidential lotteries will be discussed in Section \ref{subsec:partial}.

\subsection{Upper and Lower Expected Utilities}
\label{subsec:upperlower}

Let $\mu$ be a mass function on $\calC$, and $u$ a utility function $\calC \rightarrow \reels$. The \emph{lower} and \emph{upper expectations} of $u$ with respect to $\mu$ are defined, respectively, as the averages of the minima and  the maxima of $u$ within each focal set of $\mu$ \cite{dempster67a,shafer81,dempster87}:
\begin{subequations}
\label{eq:Einfsup}
\begin{align}
\label{eq:Einf}
\Einf_\mu(u)&=\sum_{A \subseteq \calC} \mu(A) \min_{c\in A} u(c),\\
\label{eq:Esup}
\Esup_\mu(u)&=\sum_{A \subseteq \calC} \mu(A) \max_{c\in A} u(c).
\end{align}
\end{subequations}
It is clear that $\Einf_\mu(u)\le \Esup_\mu(u)$, with the inequality becoming an equality when $\mu$ is Bayesian, in which case the lower and upper expectations  collapse to the usual expectation. If $\mu$ is logical with focal set $A$, then $\Einf_\mu(u)$ and $\Esup_\mu(u)$ are, respectively, the minimum and the maximum of $u$ in $A$. As shown in \cite{gilboa94}, the lower and upper expectations are Choquet integrals \cite{choquet53} with respect to the belief and plausibility functions, respectively. Consequently, they are consistent with the decision theories proposed by Gilboa \cite{gilboa87} and Schmeidler \cite{schmeidler89}, as discussed in Section \ref{subsec:proba}. We can also mention the axiomatic justification of the lower expectation proposed by Gilboa and Schmeidler \cite{gilboa89}.

Based on the notions of lower and upper expectations, we can define two complete preference relations among evidential lotteries as
\begin{equation}
\label{eq:pessim}
\mu_1 \pref_*  \mu_2 \text{ iff }  \Einf_{\mu_1}(u) \ge \Einf_{\mu_2}(u)
\end{equation}
and 
\begin{equation}
\label{eq:optim}
\mu_1 \pref^*  \mu_2 \text{ iff }  \Esup_{\mu_1}(u) \ge \Esup_{\mu_2}(u).
\end{equation}
Relation $\pref_*$ corresponds to a pessimistic (or conservative) attitude of the DM, since it takes in account the least favorable consequence within each focal set. When $\mu$ is logical,  $\pref_*$ corresponds to the maximin criterion; symmetrically, $\pref^*$ corresponds to an optimistic attitude and extends the maximax criterion.  For this reason, the strategies of maximizing the lower and upper expected utilities can be referred to as \emph{(generalized) maximin} and \emph{(generalized) maximax}, respectively. Both criteria boil down to the EU criterion when $\mu$ is Bayesian. In the general case, each focal set of $\mu$ corresponds to a set of possible consequences without any  probabilities assigned to them. The utility of that set is, thus, computed using one of the criteria reviewed in Section \ref{subsec:ignorance} for decision under ignorance; the utilities of the different focal sets $A$ are then weighted by their masses (probabilities) $m(A)$ and averaged in a way consistent with the notion of expected utility.

\begin{Exp}
\label{ex:invest_cred}
Consider again the investment example, with the utility matrix shown in Table \ref{tab:crit_ignorance}. Assume that uncertainty about the state of nature $\Omega$ is described by the following mass function:
\[
m(\{\omega_1\})=0.4, \quad m(\{\omega_1,\omega_2\})=0.2, \quad m(\{\omega_3\})=0.1, \quad m(\Omega)=0.3.
\]
Consider, for instance, act $f_1$. It induces the following evidential lottery:
\begin{equation*}
\label{eq:mu1}
\mu_1(\{c_{11}\})=0.4, \quad \mu_1(\{c_{11},c_{12}\})=0.2, \quad \mu_1(\{c_{13}\})=0.1, \quad \mu_1(\{c_{11},c_{12},c_{13}\})=0.3,
\end{equation*}
with $u(c_{11})=37$, $u(c_{12})=25$ and $u(c_{13})=23$. Consequently, the lower and upper expected utilities can be computed as
\begin{align*}
\Einf_{\mu_1}(u)&= 0.4\times 37 + 0.2 \times 25 + 0.1\times 23+0.3\times 23= 29\\
\Esup_{\mu_1}(u)&= 0.4\times 37 + 0.2 \times 37 + 0.1\times 23+0.3\times 37=35.6
\end{align*}
The lower and upper expected expectations for the fours acts are shown in Table \ref{tab:lower_upper}. The corresponding strict preference relations among acts are
$
f_2 \succ_* f_1 \succ_* f_4 \succ_* f_3
$
and
$
f_2 \succ^* f_3 \succ^* f_4 \succ^* f_1.
$
\end{Exp}

\begin{table}
\begin{center}
\caption{Lower and upper expected utilities for  the payoff matrix of Example \ref{ex:invest} and the mass function on $\Omega$ defined in Example \ref{ex:invest_cred}. \label{tab:lower_upper}}
\begin{tabular}{cccccc}
\hline
 & $u_{i1}$ & $u_{i2}$ & $u_{i3}$ & $\Einf_{\mu_i}(u)$ & $\Esup_{\mu_i}(u)$\\
\hline
$f_1$ & 37 & 25 &23 & 29.0 &35.6 \\
$f_2$ & 49 & 70 &2&\textbf{30.2} & \textbf{54.8}\\
$f_3$ & 4 & 96 &1&2.8& 49.7\\
$f_4$ &22 & 76 &25&22.3& 49.3\\
\hline
\end{tabular}
\end{center}
\end{table}

\subsection{Generalized Hurwicz Criterion}
\label{subsec:hurwicz}

Just as the lower and upper expected utility models generalize, respectively, the maximin and  maximax criteria, the Hurwicz criterion (\ref{eq:hurwicz}) can be readily generalized by defining the expectation of $u$, for a pessimism index $\alpha \in [0,1]$, as
\begin{subequations}
\label{eq:hurwicz_bf}
\begin{align}
\esp_{\mu,\alpha}(u) &= \sum_{A \subseteq \calC} \mu(A) \left( \alpha \min_{c\in A} u(c) + (1-\alpha) \max_{c\in A} u(c) \right) \label{eq:hurwicz_bf1}\\
&= \alpha \Einf_\mu(u) + (1-\alpha) \Esup_\mu(u).
\end{align}
\end{subequations}
A more general version of this criterion (where $\alpha$ in (\ref{eq:hurwicz_bf1}) depends on $A$) was first introduced by Jaffray \cite{jaffray88,jaffray89}, who also  justified it axiomatically (see Section \ref{subsec:axiom} below). Criterion (\ref{eq:hurwicz_bf}) with fixed $\alpha$ was later discussed by Strat \cite{strat90}, who proposed to interpret $\alpha$ as the DM's subjective probability that the ambiguity will be resolved unfavorably (see also \cite{srivastava97} for a discussion of this criterion).  Hereafter, we will use the term ``Hurwicz criterion'' to refer to decision based on (\ref{eq:hurwicz_bf}), as this principle is a direct extension to the Hurwicz criterion in the case of complete ignorance.

Recently, Ma et al. \cite{ma17} proposed to determine $\alpha$ automatically as a function of $\mu$ by equating it with the normalized nonspecificity measure \cite{dubois85b,ramer87} of $\mu$ defined as
\[
N(\mu)=\frac{1}{\log_2 \vert \calC\vert} \sum_{A \subseteq \calC} \mu(A)\log_2(A).
\]
Seeing $N(\mu)$ as a measure of ambiguity, this approach corresponds to a decision model  in which the DM is all the more cautious that there is more ambiguity. Ma et al. show that this model is subject neither to Ellsberg's paradox \cite{ellsberg61}, nor to a more recent paradox proposed by Machina \cite{machina09}, which contradicts Choquet expected utility. However, this decision criterion seems to be, otherwise, weakly justified.

\subsection{Pignistic Criterion}
\label{subsec:pignistic}

A completely different approach to decision-making with belief function was advocated by Smets as part of the \emph{Transferable Belief Model}, a variant of  DS theory  \cite{smets90e,smets94a,smets02b}. Smets defended a two-level mental model, composed of a \emph{credal level}, where an agent's belief are represented by belief functions, and the \emph{pignistic level}, where decisions are made by maximizing EU with respect to a probability measure derived from a belief function.  The rationale for introducing probabilities at the decision level is the avoidance of Dutch books, i.e., sequences of bets than incur sure loss \cite{vineberg16}. Furthermore, Smets \cite{smets05b} argued  that, as the consequence of the MEU principle, the  belief-probability transformation $T$ should be linear, i.e., it should verify
\begin{equation}
\label{eq:linear_pignistic}
T\left(\alpha \mu_1 + (1-\alpha) \mu_2\right)=\alpha T(\mu_1) + (1-\alpha) T(\mu_2),
\end{equation}
for any mass functions $\mu_1$ and $\mu_2$ and for any $\alpha\in [0,1]$. He then showed that the only transformation $T$ verifying (\ref{eq:linear_pignistic}) is the \emph{pignistic transformation}, with $p_\mu=T(\mu)$  given by
\begin{equation}
p_\mu(c) = \sum_{A\subseteq \calC } \frac{\mu(A)}{|A|} I(c\in A),
\end{equation}
for any $c\in \calC$. The pignistic probability $p_\mu$ turns out to be  mathematically identical to the  Shapley value in cooperative game theory \cite{shapley53}. The expected utility w.r.t. the pignistic probability is
\begin{eqs}
\esp_{p}(u) &= \sum_{c \in \calC} p_\mu(c) u(c)\\
                  &= \sum_{c \in \calC}  u(c) \sum_{\{A\subseteq \calC | c\in A\}} \frac{\mu(A)}{|A|}\\
                  &=\sum_{\{A\subseteq \calC\}} \mu(A) \left(\frac{1}{|A|} \sum_{c\in A} u(c)\right).
\end{eqs}
The maximum pignistic expected utility criterion thus averages the mean utility inside each focal set $A$. Consequently, it extends the Laplace criterion discussed in Section \ref{subsec:ignorance}, when uncertainty is quantified by a belief function. 

\begin{Exp}
\label{ex:pig}
Continuing Example \ref{ex:invest_cred}, the pignistic expected utility of act $f_1$ is
\[
\esp_{p_1}(u)=0.4\times 37 + 0.2\times\frac{37+25}{2} + 0.1\times 23+ 0.3\times\frac{37+25+23}{3}=31.8.
\]
Similarly, we find  $\esp_{p_2}(u)=43.8$, $\esp_{p_3}(u)=21.8$ and $\esp_{p_4}(u)=33.4$.
\end{Exp}

\begin{Rem}
We have seen in Section \ref{subsec:ignorance} that the Laplace criterion for decision-making under ignorance may lead to different decisions when a state of nature is ``duplicated'', i.e., when the state space $\Omega$ is refined. The pignistic criterion obviously has the same drawback: refining the frame of discernment  changes the pignistic probability. Smets \cite{smets94a} tried to circumvent this difficulty by stating that the DM needs to select a ``betting frame'' before computing the pignistic probability. It is not always clear, however, on which basis such a choice can be made. Wilson \cite{wilson93} showed that the lower and upper expectations (\ref{eq:Einfsup}) are, respectively, the minimum and the maximum of the pignistic expectations  computed over all refinements of the frame of discernment.  
\end{Rem}

\begin{Rem}
There are obviously other ways of transforming a belief function into a probability distribution. Voorbraak \cite{voorbraak89} and Cobb and Shenoy \cite{cobb06} have argued for the \emph{plausibility transformation}, which approximates a belief function by a probability distribution, in such a way that the probability of singletons is proportional to their plausibility. This transformation has the remarkable property of being compatible with Dempster's rule (i.e., the approximation of the orthogonal sum of two belief functions is the orthogonal sum of their approximations). This property makes the plausibility transformation suitable for approximating a DS model by a probabilistic model.   To our knowledge, no  argument has been put forward in favor of using this approximating probability distribution for decision-making. 
\end{Rem}

\begin{Rem}
Although the pignistic criterion does not seem to depend on any parameter (which may be key to its appeal in real applications), it does depend on the granularity level of the frame of discernment. Smets was aware of this difficulty and assumed that a ``betting frame'' had been chosen prior to decision-making. While this choice may be natural in some applications, this may not always be the case. Wilson \cite{wilson93} showed that, when considering all refinements of the current frame, the pignistic expectation ranges between the lower and upper expectations, just as the generalized Hurwicz criterion does. When using the pignistic criterion, the choice of a betting frame is, thus, a critical design issue, which is often overlooked. 
\end{Rem}

\subsection{Generalized OWA Criterion}

A more general family of expected utility criteria can be defined by aggregating the utilities $u(c)$ within each focal set $A \subseteq \calC$ using OWA operators  as recalled in Section \ref{subsec:ignorance} \cite{yager92}. It is clear that the previous definitions in  Sections \ref{subsec:upperlower}, \ref{subsec:hurwicz} and \ref{subsec:pignistic} are recovered as special cases. To determine the weights of the  OWA operators, Yager \cite{yager92} proposed to fix the degree of optimism $\beta$ defined by (\ref{eq:optimism}), and to use the maximum-entropy operators, for each cardinality $|A|$. Formally,
\begin{equation}
\esp^{\textrm{owa}}_{\mu,\beta}(u) = \sum_{A\subseteq \calC} \mu(A) F_{|A|,\beta}(\{u(c) | c\in A\}),
\end{equation}
where $F_{|A|,\beta}$ is the maximum-entropy OWA operator with degree of optimism $\beta$ and arity $|A|$. We can remark that parameter $\beta$ plays the same role here, and has roughly the same interpretation, as one minus the pessimism index $\alpha$ in the Hurwicz criterion. However, each quantity $F_{|A|,\beta}(\{u(c) | c\in A\})$ depends on all the values $u(c)$ for all $c\in A$, and not only on the minimum and the maximum, and the pignistic criterion is recovered for $\beta=0.5$. This method is further discussed in Ref. \cite{xiong14}.

\begin{Exp}
\label{ex:gen_OWA}
Considering again the investment example and the mass function of Examples \ref{ex:invest_cred} and \ref{ex:pig}, Figure \ref{fig:gen_hurwicz_OWA} shows the aggregated utilities for the Hurwicz criteria as functions of $\alpha$ (Figure \ref{fig:gen_hurwicz}) and for the generalized OWA criterion as functions of $1-\beta$ (Figure \ref{fig:gen_OWA}). Once again, we can see that these two criteria yield similar results, and that the pignistic expectations are obtained as a special case of the generalized OWA criterion with $\beta=0.5$.

\begin{figure}
\centering 
\centering 
\subfigure[\label{fig:gen_hurwicz}]{\includegraphics[width=0.49\textwidth]{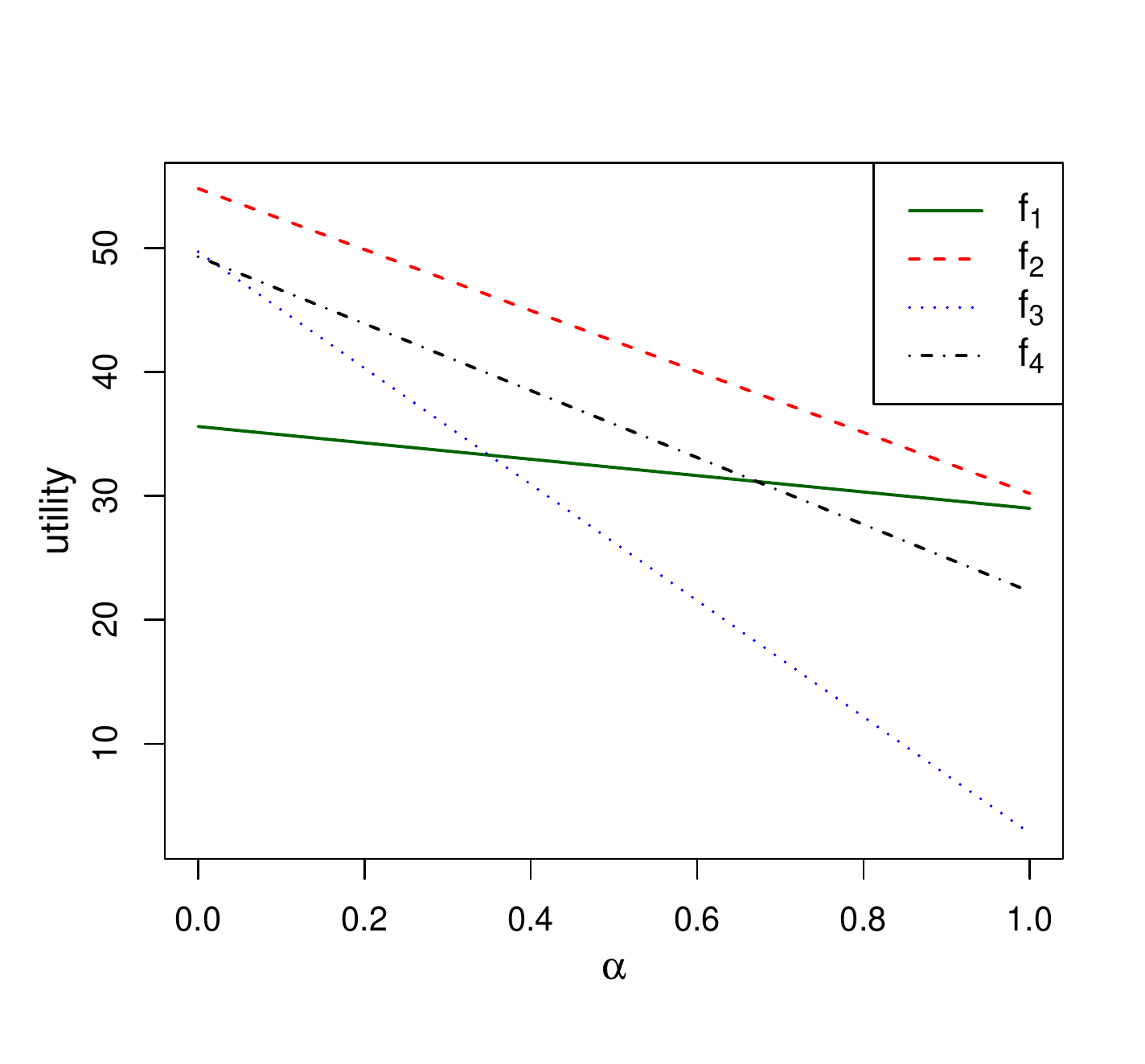}}
\subfigure[\label{fig:gen_OWA}]{\includegraphics[width=0.49\textwidth]{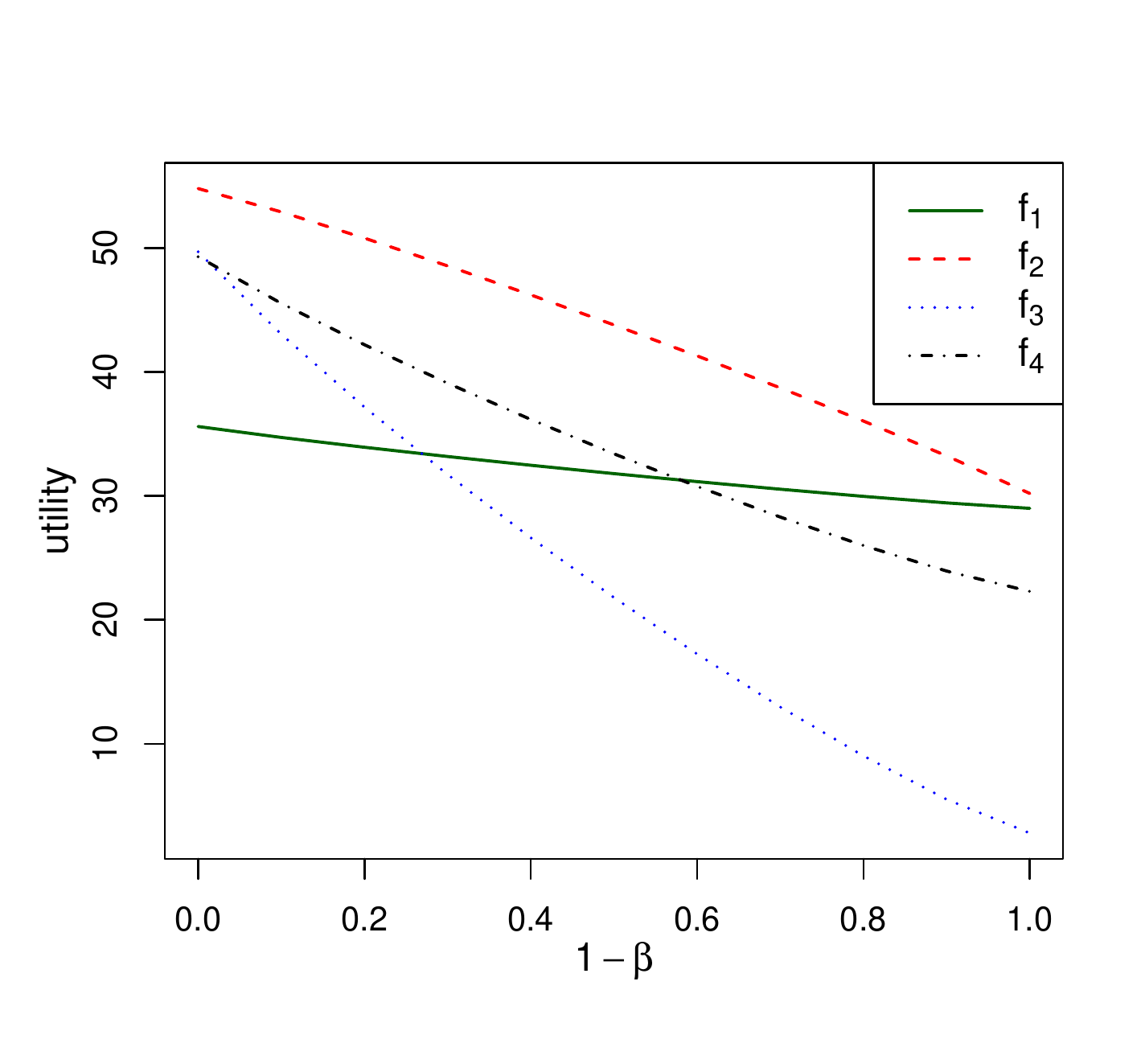}}
\caption{Aggregated utilities vs. pessimism index $\alpha$ for the generalized Hurwicz criterion (a) and vs. one minus the degree of optimism $\beta$ for the generalized OWA criterion (b).  (Example  \ref{ex:gen_OWA}). \label{fig:gen_hurwicz_OWA}}
\end{figure}
\end{Exp}

\subsection{Generalized Minimax Regret}
\label{subsec:gen_minimax_regret}

Finally, Yager \cite{yager04a} also extended the minimax regret criterion  to belief functions. As in  Section \ref{subsec:ignorance}, we need to consider $n$ acts $f_1,\ldots,f_n$, and we  write $u_{ij}=u[f_i(\omega_j)]$.  The regret if act $f_i$ is selected, and state $\omega_j$ occurs, is $r_{ij}=\max_k u_{kj} - u_{ij}$. For a non-empty subset $A$ of $\Omega$, the maximum regret of act $f_i$ is 
\begin{equation}
R_i(A)=\max_{\omega_j\in A} r_{ij}.
\end{equation}
Given a mass function $m$ on $\Omega$, the \emph{expected maximal regret} for act $f_i$ is
\begin{equation}
\overline{R}_i=\sum_{\emptyset\neq A\subseteq \Omega} m(A)R_i(A).
\end{equation}
Using the \emph{generalized minimax regret} criterion, act $f_i$ is preferred over act $f_k$ if $\overline{R}_i \le \overline{R}_k$. The minimax regret criterion of decision-making under ignorance is recovered when $m$ is logical. If $m$ is Bayesian, we have
\begin{eqs}
\overline{R}_i&=\sum_{j} m(\{\omega_j\}) r_{ij}\\
                     &=\sum_{j} m(\{\omega_j\}) (\max_k u_{kj} - u_{ij})\\
                     &= \sum_{j} m(\{\omega_j\}) \max_k u_{kj} - \sum_{j} m(\{\omega_j\})u_{ij} . \label{eq:last}
\end{eqs}
The first term on the right-hand side of Eq. (\ref{eq:last}) does not depend on $i$, and the second one is the expected utility. Hence, the generalized minimax regret criterion is identical to the MEU model when $m$ is Bayesian.

\begin{Exp}
The regrets $r_{ij}$ for  the investment data are given in Table \ref{tab:maxregret}. With the mass function of Example  \ref{ex:invest_cred}, we get, for act $f_1$:
\[
\overline{R}_1=0.4\times 12 + 0.2\times 71 +0.1\times 2 + 0.3\times 71=40.5.
\]
Similarly, we have $\overline{R}_2=15.3$, $\overline{R}_3=42.9$ and $\overline{R}_4=24.3$. The corresponding preference relation is, thus, $f_2\succ f_4\succ f_1 \succ f_3$.
\end{Exp}

\subsection{Jaffray's and related axioms}
\label{subsec:axiom}

Except for  generalized minimax regret, the criteria for decision-making with belief functions decision  reviewed   above are all of the form
\begin{equation}
\mu_1 \succcurlyeq \mu_2 \text{ iff } U(\mu_1) \ge U(\mu_2),
\end{equation}
where $U$ is a function from the set of evidential lotteries to $\reels$, such that
\begin{equation}
\label{eq:U}
U(\mu)=\sum_{\emptyset\neq A\subseteq \calC} \mu(A) U(\mu_A),
\end{equation}
where $\mu_A$ is the logical mass function with focal set $A$. To simplify the notation, we can write $U(A)$ in place of $U(\mu_A)$, and $u(c)$ for $U(\{c\})$. With these notations, we have
\bi
\item $U(A)=\min_{c\in A} u(c)$ for the maximin criterion;
\item $U(A)=\max_{c\in A} u(c)$ for the maximax criterion;
\item $U(A)=\alpha \min_{c\in A} u(c) + (1-\alpha) \max_{c\in A} u(c)$ for the Hurwicz criterion;
\item $U(A)=(1/|A|)\sum_{c\in A} u(c)$ for the pignistic  criterion;
\item $U(A)=F_{|A|,\beta}(\{u(c) | c\in A\})$ for the OWA  criterion.
\ei
Jaffray \cite{jaffray89} showed that a preference relation $\pref$ among evidential lotteries is representable by a linear utility function verifying (\ref{eq:U}) if and only if it verifies the Von Neumann and Morgenstern axioms (see Section  \ref{subsec:proba}) extended to evidential lotteries, i.e.,
\bd
\item[Transitivity and Completeness:] $\pref$ is a transitive and complete relation (i.e., a complete preorder);
\item[Continuity:] for all $\mu_1$, $\mu_2$ and $\mu_3$  such that $\mu_1 \succ \mu_2 \succ \mu_3$, there exists $\alpha$, $\beta$ in $(0, 1)$ such that 
\begin{equation}
\alpha \mu_1+ (1- \alpha)\mu_3 \succ \mu_2 \succ \beta \mu_1+ (1- \beta)\mu_3;
\end{equation}
\item[Independence:] for all $\mu_1$, $\mu_2$ and $\mu_3$, and for all $\alpha$ in (0, 1), $\mu_1 \succ \mu_2$ implies
\begin{equation}
\alpha \mu_1+ (1 - \alpha)\mu_3 \succ \alpha \mu_2+ (1 - \alpha)\mu_3.
\end{equation}
\ed
It is clear that $U(\mu)$ in (\ref{eq:U}) becomes the expected utility when $\mu$ is Bayesian: we then have
\begin{equation}
\label{eq:Ubayes}
U(\mu)=\sum_{c\in  \calC} \mu(\{c\}) u(c).
\end{equation}
The major difference with the classical EU model is that  we now need, in the general case, to elicit the utility values $U(A)$ for each subset $A\subseteq \calC$ of consequences, which limits the practical use of the method. However, Jaffray \cite{jaffray89} showed that a major simplification of (\ref{eq:U}) can be achieved by introducing an additional axiom. To present this axiom, let us introduce the following notation. Let us write $c_1 \pref c_2$ whenever $\mu_{\{c_1\}} \pref \mu_{\{c_2\}}$. Furthermore, let $\cinf_A$ and $\csup_A$ denote, respectively, the worst and the best consequence in $A$. The additional axiom can then be stated as follows:

\bd
\item[Dominance:] for all non-empty subsets $A$ and $B$ of $\calC$, if $\cinf_A \pref \cinf_B$ and  $\csup_A \pref \csup_B$, then $\mu_A \pref \mu_B$.
\ed

This axiom was justified by Jaffray \cite{jaffray89} as follows. If  $\cinf_A \pref \cinf_B$ and  $\csup_A \pref \csup_B$, it is possible to construct a set $\Omega$ of states of nature, and two acts $f:\Omega\rightarrow A$ and $f':\Omega\rightarrow B$, such that, for any $\omega\in\Omega$, $f(\omega)\pref f'(\omega)$. As act $f$ dominates $f'$, it should be preferred whatever the information on $\Omega$. Hence, $f$ should be preferred to $f'$ when we have a vacuous mass function on $\Omega$, in which case $f$ and $f'$ induce, respectively, the logical mass function $\mu_A$ and $\mu_B$ on $\calC$.

The Dominance axiom immediately implies that, for any non-empty subsets $A$ and $B$ of $\calC$, if $\cinf_A \sim \cinf_B$ and  $\csup_A \sim \csup_B$, then $\mu_A \sim \mu_B$, and $U(A)=U(B)$. Hence, $U(A)$ depends only on the worst and the least consequence in $A$, and we can write $U(A)=u(\cinf_A,\csup_A)$. Equation (\ref{eq:U}) thus becomes
\begin{equation}
\label{eq:U1}
U(\mu)=\sum_{\emptyset\neq A\subseteq \calC} \mu(A) u(\cinf_A,\csup_A).
\end{equation}
If one accepts the Dominance axiom, one is then led to rejecting the pignistic criterion, as well as the generalized OWA criterion, except when it is identical to the Hurwicz criterion. 

To describe the DM's attitude to ambiguity, Jaffray \cite{jaffray89} then introduced the \emph{local pessimism index} $\alpha(\cinf,\csup)$, defined as  the value of $\alpha$ which makes the DM indifferent between: 
\be
\item  Receiving at least $\cinf$ and at most $\csup$, with no further information, and
\item  Receiving either $\cinf$ with probability $\alpha$ or $\csup$ with probability $1-\alpha$.
\ee
We then have
\begin{equation}
u(\cinf,\csup)=\alpha(\cinf,\csup) u(\cinf) + (1-\alpha(\cinf,\csup))u(\csup).
\end{equation}
This relation shows how the DM's attitudes to risk and to ambiguity jointly determine $U$. Now, the utility of evidential lottery $\mu$ can be written as
\begin{equation}
\label{eq:U2}
U(\mu)=\sum_{\emptyset\neq A\subseteq \calC} \mu(A) \left[\alpha(\cinf_A,\csup_A) u(\cinf_A)\right. + 
\left. (1-\alpha(\cinf_A,\csup_A))u(\csup_A)\right].
\end{equation}
The Hurwicz criterion (\ref{eq:hurwicz_bf}) corresponds to the case where $\alpha(\cinf,\csup)$ is equal to a constant $\alpha$.

Jaffray's axioms are the counterpart of the axioms of Von Neumann and Morgenstern (see Section \ref {subsec:proba}) for decision under risk: assuming uncertainty about the consequences of each act to be described by  belief functions, they justify the decision strategy  maximizing the utility criterion (\ref{eq:U2}) for  evidential lotteries. 

Several researchers have arrived at (\ref{eq:U2}) from different sets of axioms. Jaffray and Wakker \cite{jaffray94b} (see also \cite{wakker00}) consider the situation where objective probabilities are defined on a finite set $S$, and there is a multi-valued mapping $\Gamma$ that maps each element $s\in S$ to a subset $\Gamma(s)$ of the set $\Omega$ of states of nature. The authors thus do not postulate a belief function in the first place, but started one step before, i.e., they postulate the existence of a source, which induces a belief function $Bel$ (see Section \ref{subsec:rec_DS}). Each act $f:\Omega\rightarrow \calC$ then carries $Bel$ from $\Omega$ to $\calC$, i.e., it induces an evidential lottery. The authors then justify a  neutrality axiom, which states that two  acts are indifferent whenever they generate the same evidential lottery. Finally, they derive the decision criterion (\ref{eq:U2}) from two axioms: a continuity condition, and a weakened version of Savage's sure-thing principle.  Interestingly, a similar criterion was obtained in \cite{ghirardato01}, for the case where an act is defined as a multi-valued mapping from $\Omega$ to $2^\calC$. By postulating axioms similar to those of Savage, and two additional axioms,  Ghirardato proved that the preference relation among acts is represented by a utility function similar to (\ref{eq:U2}). Finally, Zhou et al. \cite{zhou18} recently managed to provide a set of axioms that guarantee  the existence of a utility function and a unique belief function on the state space $\Omega$ such that preferences among acts are governed by (\ref{eq:U2}). Although the axioms are quite technical and their meaning may be difficult to grasp, this result seems to be the closest so far to a belief-function counterpart of Savage's theorem.

For completeness and to conclude this section, we should also mention  an alternative set of axioms proposed by Giang and Shenoy \cite{giang11}, leading to a different decision criterion (see also \cite{giang12}). Their approach, however, is restricted to the case of partially consonant mass functions. A mass function $m$ is said to be partially consonant if  its focal sets can be divided into groups such that (a) the focal sets of different groups do not intersect and (b) the focal sets of the same group are nested. The family of partially consonant mass functions includes Bayesian and consonant mass functions as special cases. In the context of statistical inference, Walley \cite{walley87} has shown that partially consonant mass functions arise as a consequence of some axioms. There does not seem, however, to be any compelling reason for constraining belief functions to be partially consonant outside the specific context of statistical inference.  

\subsection{Dropping the Completeness Requirement}
\label{subsec:partial}

All decision criteria reviewed in Sections \ref{subsec:upperlower} to \ref{subsec:gen_minimax_regret} induce a complete preference relation of acts. In some applications, however, the necessity of the completeness requirement may be questioned, and it can be argued that the potential ``indecisiveness'' of the agents should be allowed \cite{dubra04}. Hereafter, we review two categories of decision criteria inducing partial preferences relations based, respectively, on lower/upper expectations and on extensions of  stochastic dominance.

\subsubsection*{Criteria based on lower and upper expectations}

If one drops the requirement that the preference relation among evidential lotteries be complete, then one can adopt the following partial preference relation, called the \emph{strong dominance} or \emph{interval dominance} relation \cite{troffaes07}:
\begin{equation}
\label{eq:ID}
\mu_1 \pref_{SD} \mu_2 \text{ iff } \Einf_{\mu_1}(u) \ge \Esup_{\mu_2}(u).
\end{equation}
Given a collection of evidential lotteries $\mu_1,\ldots,\mu_n$, we can then consider the set of non-dominated elements with respect to the strict preference relation $\succ_{SD}$. The choice set is then
\begin{equation}
\label{eq:set_ID}
\calM_{SD} = \{\mu \in \{\mu_1,\ldots,\mu_n\} \mid \forall \mu' \in \{\mu_1,\ldots,\mu_n\}, \neg(\mu' \succ_{SD} \mu)\}.
\end{equation}
However, condition (\ref{eq:ID}) is very strong, and many pairs of mass functions will typically not be comparable. As a consequence,  choice set (\ref{eq:set_ID}) will often be too large. Additionally, the strong dominance relation (\ref{eq:ID}) seems hard to justify outside the imprecise-probability setting (see Section \ref{sec:IP}).

A weaker partial preference relation is the \emph{interval bound dominance} relation \cite{destercke10} defined as follows:
\begin{equation}
\label{eq:WD}
\mu_1 \pref_{ID} \mu_2 \text{ iff } \left(\Einf_{\mu_1}(u) \ge \Einf_{\mu_2}(u)\right) \text{ and } \left(\Esup_{\mu_1}(u) \ge \Esup_{\mu_2}(u)\right).
\end{equation}
Obviously, $\mu_1 \pref_{SD} \mu_2$ implies $\mu_1 \pref_{ID} \mu_2$, so that relation $\pref_{ID}$ compares more pairs of mass functions than $\pref_{SD}$ does. Further, interval bound dominance can be justified as follows: $\mu_1$ is at least as desirable as $\mu_2$ according to interval bound dominance iff it is at least as desirable as $\mu_2$ according to the Hurwicz criterion for any value of the pessimism index $\alpha$. Formally,
\begin{equation}
\left(\mu_1 \pref_{ID} \mu_2\right) \Leftrightarrow \left(\forall \alpha \in [0,1], \; \esp_{\mu_1,\alpha}(u) \ge  \esp_{\mu_2,\alpha}(u)\right),
\end{equation}
where $\esp_{\mu_1,\alpha}(u)$ and $\esp_{\mu_2,\alpha}(u)$ are defined by Eq. (\ref{eq:hurwicz_bf}). interval bound dominance thus corresponds to a conservative approach that seems appropriate if one accepts the Hurwicz criterion while being totally ignorant about the DM's attitude towards ambiguity.

\begin{Exp}
As we can see from Table \ref{tab:lower_upper} and Figure \ref{fig:gen_hurwicz_OWA}, no pair of acts $(f_i,f_j)$ is such that $\Einf_{\mu_i}(u) \ge \Esup_{\mu_j}(u)$. Consequently, the choice set for the strong dominance relation is $\{f_1,f_2,f_3,f_4\}$, i.e., the strong dominance criterion does not allow us to compare any of the four acts. For the interval bound dominance criterion, we can see that act $f_2$ dominates the other three acts, but is not dominated by any other act. Consequently, the choice set for the interval bound dominance relation is $\{f_2\}$.
\end{Exp}

\subsubsection*{Criteria based on extensions of  stochastic dominance } 

As shown in  \cite{denoeux09b}, the notion of first-order dominance can be generalized to belief functions on the real-line\footnote{See also \cite{montes14a} and \cite{jansen18b} for further generalizations in the imprecise probability setting}. Specifically, consider a gamble $X=u\circ f$ induced by an act $f$. If $\mu_f$ denotes the associated evidential lottery, then the uncertainty on $X$ is described by the mass function $m_X$ obtain by carrying $\mu_f$ to $\reels$ by mapping $u$, i.e.,
\[
m_X(A)=\sum_{\{B\subseteq \calC \mid u[B]=A\}} \mu_f(B).
\]
Equation (\ref{eq:StochD}) can be extended in several ways to define the preference between mass functions $m_X$ and $m_Y$ induced by gambles $X$ and $Y$:
\[
 \begin{array}{cclc}
 m_X \gew m_Y & \Longleftrightarrow & \forall x \in \reels, & Pl_X((x,+\infty)) \ge Bel_Y((x,+\infty)) \\
 m_X \geinf m_Y &  \Longleftrightarrow & \forall x \in \reels, &Bel_X((x,+\infty)) \ge Bel_Y((x,+\infty)) \\
 m_X \gesup m_Y  &  \Longleftrightarrow & \forall x \in \reels, &Pl_X((x,+\infty)) \ge Pl_Y((x,+\infty)) \\
 m_X \gedr m_Y  &  \Longleftrightarrow & \forall x \in \reels, &Bel_X((x,+\infty)) \ge Pel_Y((x,+\infty))
\end{array}
\]
Obviously, all four relations boil down to the first-order stochastic dominance relation when the evidential lotteries are Bayesian. These four credal ordering relations have the following properties \cite{denoeux09b}, which parallel the property of stochastic dominance with respect to expectation:
\[
 \begin{array}{cccl}
 m_X \gew m_Y & \Longleftrightarrow & \forall h \in \calH, & \Esup[h(X)] \ge \Einf[h(Y)]\\
 m_X \geinf m_Y &  \Longleftrightarrow & \forall h \in \calH, & \Einf[h(X)] \ge \Einf[h(Y)] \\
 m_X \gesup m_Y  &  \Longleftrightarrow & \forall h \in \calH, & \Esup[h(X)] \ge \Esup[h(Y)] \\
 m_X \gedr m_Y  &  \Longleftrightarrow & \forall h \in \calH, & \Einf[h(X)] \ge \Esup[h(Y)],
\end{array}
\]
where $\calH$ denotes the set of bounded and non decreasing functions from $\reels$ to $\reels$. Consequently, relations $\geinf$, $\gesup$ and $\gedr$ correspond, respectively, to the maximin, maximax and strong dominance criteria when the utility function is only defined up to a non decreasing transformation.

Finally, we can remark that statistical preference has been extended in the imprecise probability setting by Montes et al. \cite{montes14b} and Couso et al. \cite{couso15}. Extensions in the DS setting remain to be explored.

\section{Imprecise-Probability View}
\label{sec:IP}

As recalled in Section \ref{subsec:rec_DS}, a belief function is a coherent lower probability for  a convex set of compatible probability measures. Consequently, decision criteria proposed in the imprecise-probability framework \cite{troffaes07, huntley14} are also applicable when uncertainty is described by belief functions. To keep the exposition simple, we assume that we are in the case where we have a mass function $m$ on $\Omega$ and acts are mappings from $\Omega$ to $\calC$. For each act $f_i$, we denote by $X_i=u \circ f_i$ the gamble that maps each $\omega_j$ in $\Omega$ to the utility $u_{ij}=u[f_i(\omega_j)]$. 
The lower and upper expectations defined by Eq. (\ref{eq:Einfsup}) can be rewritten using (\ref{eq:decm1}) as follows:
\begin{eqs}
\Einf_\mu(u) &= \sum_{A \subseteq \calC} \mu(A) \inf_{c\in A} u(c)\\
&=\sum_{A \subseteq \calC} \left\{\sum_{B\subseteq \Omega: f(B)=A} m(B)\right\} \inf_{c\in A} u(c)\\
&=\sum_{B\subseteq \Omega} m(B)  \inf_{c\in f(B)} u(c)\\
&=\sum_{B\subseteq \Omega} m(B)  \inf_{\omega\in B} u(f(\omega))=\Einf_m(X), \label{eq:minmax}
\end{eqs}
with $X= u \circ f$. Similarly, $\Esup_\mu(u)$ can be written as $\Esup_m(X)$. The mappings $X \rightarrow \Einf_m(X)$ and $X \rightarrow \Esup_m(X)$ are called, respectively, \emph{lower} and \emph{upper previsions} \cite{walley91}. An important result is that  these lower and upper expectations can be interpreted as lower and upper bounds of expectations with respect to compatible probability measures   \cite{shafer81,wasserman90,gilboa94}. In other words, the mean of minima in (\ref{eq:minmax}) is also the minimum of means (expectations) with respect to all compatible probability measures; similarly, the mean of maxima  is the maximum of means \cite{gilboa94}.  
Formally, we have the following equalities:
\begin{subequations}
\label{eq:lowerupperIP}
\begin{align}
\Einf_m(X) &= \min_{P\in \calP(m)} \esp_P(X)\\
\Esup_m(X) &= \max_{P\in \calP(m)} \esp_P(X).
\end{align}
\end{subequations}
The interval $\left[\Einf_m(X),\Esup_m(X)\right]$ can, thus, be seen as the range of $\esp_P(X)$ for all probability measures $P$ in the credal set  of $m$. 

As a  consequence of the above result, the strong dominance relation (\ref{eq:ID})   has a natural interpretation in the imprecise-probability setting. Let $X_1$ and $X_2$ be two gambles. By abuse of notation, we can use the same symbol for the preference relations among gambles and among evidential lotteries. Thus,  $X_1 \pref_{SD} X_2$ iff  $\Einf_m(X_1) \ge \Esup_m(X_2)$, i.e., iff for any two probability measures $P_1$ and $P_2$ compatible with $m$,  the expectation of $X_1$ with respect to $P_1$   is always higher that the expectation of $X_2$ with respect to $P_2$. Formally,
\begin{equation}
\label{eq:SD}
X_1 \pref_{SD} X_2 \Leftrightarrow \left(\forall (P_1,P_2) \in \calP(m)^2,  \; \esp_{P_1}(X_1) \ge \esp_{P_2}(X_2)\right).
\end{equation}
A gamble $X$ is, thus, a maximal element of $\pref_{SD}$ if, for any gamble $X'$, there exists  probability measures $P$ and $Q$ in $\calP(m)$ such that $\esp_{P}(X) \ge \esp_{Q}(X')$. This set is arguably too large, as $P$ and $Q$ are not required to be  identical. Two more useful decision criteria developed  in the imprecise-probability framework  will now be discussed in Section \ref{subsec:maximality} and \ref{subsec:e-admiss}.

\subsection{Maximality}
\label{subsec:maximality} 

The \emph{maximality} criterion was introduced by Walley \cite[Section 3.9, page 160]{walley91} (see also \cite{walley00,couso11b,miranda08}). It states that gamble $X_1$ is ``almost as desirable''  as $X_2$ if the lower expectation of $X_1-X_2$ is positive:
\begin{equation}
\label{eq:maximality}
X_1 \pref_{max} X_2 \text{ iff } \Einf_m(X_1-X_2) \ge 0,
\end{equation}
and the preference is strict if the inequality in the right-hand side of  (\ref{eq:maximality}) is strict. In Walley's theory, a lower prevision is interpreted as the highest price an agent is willing to pay to acquire a gamble. Clearly, Eq. (\ref{eq:maximality}) expresses that the DM is willing to pay a positive price to get $X_1$ instead of $X_2$, i.e., that  he strictly prefers $X_1$ to $X_2$. In terms of credal set, the condition $\Einf_m(X_1-X_2) \ge 0$ means that, for any compatible probability $P$, the expectation of $X_1-X_2$ with respect to $P$ is positive, i.e.:
\begin{equation}
\label{eq:maximality2}
X_1 \pref_{max} X_2 \text{ iff } \left(\forall P \in \calP(m), \; \esp_P(X_1) \ge \esp_P(X_2)\right).
\end{equation}
Comparing Eqs (\ref{eq:SD}) and (\ref{eq:maximality2}), it is clear that $X_1 \pref_{SD} X_2 \Rightarrow X_1 \pref_{max} X_2$, and the implication is strict. A gamble $X$ is a maximal element of $\succ_{max}$ iff, for any gamble $X'$, there exists $P \in \calP(m)$ such that $\esp_{P}(X) \ge \esp_{P}(X')$. The set of maximal elements of $ \succ_{max}$ is, thus, included in that of $\succ_{SD}$.

The maximality criterion thus seems to be better founded  and more useful than strong dominance. However, these advantages come at a price, as finding the maximal elements according to the maximality criterion requires computing $n^2-n$ lower expectations  (where $n$ is the number of gambles), against $n$ lower expectations and as many upper expectations for the strong dominance criterion \cite{troffaes07}. Hence, strong dominance has a computational advantage when the number of alternatives is large. As the choice set of $\succ_{SD}$ contains that of $\succ_{max}$, it can also be computed as a preliminary step to reduce the number operations needed to implement the maximality criterion.

Finally, we can remark that maximality,  although introduced and studied in the imprecise-probability context, also makes sense regardless of any notion of imprecise probability. In the DS framework, Eq.  (\ref{eq:maximality}) can be understood to mean that, if one DM selects $X_1$ and another DM selects $X_2$, then the former is expected to gain a higher utility, even assuming that ambiguity will be resolved in favor of the latter.

\begin{Exp}
\label{ex:maximality}
Consider again the pay-off matrix of Table \ref{tab:crit_ignorance} and the mass function $m$ of Example \ref{ex:invest_cred}. We have $X_1(\omega_1)=37$, $X_1(\omega_2)=25$, $X_1(\omega_3)=23$ and $X_2(\omega_1)=49$, $X_2(\omega_2)=70$, $X_2(\omega_3)=2$. Hence,
\[
\Einf_m(X_1-X_2) = 0.4\times (-12) + 0.2\times (-45)+ 0.1\times 21 + 0.3\times (-45)= -25.2
\]
and
\[
\Einf_m(X_2-X_1) = 0.4\times (12) + 0.2\times (12)+ 0.1\times (-21) + 0.3\times (-21)= -1.2.
\]
Consequently,  $X_1$ and $X_2$ are not comparable by the maximality criterion. The matrix $\Delta X$ with general term $[\Delta X]_{ij}=\Einf_m(X_i-X_j)$ is
\[
\Delta X=\begin{bmatrix}
  \cdot &-25.2& -20.1& -19.7\\
 -1.2 &  \cdot &  5.1&   0.4\\
 -31.9 &-40.6  & \cdot &-20.4\\
-13.3 &-22.0 & -0.4 &  \cdot 
\end{bmatrix}.
\]
We thus have $X_2 \succ_{max} X_3$ and $X_2 \succ_{max} X_4$, whereas $X_1$ and $X_2$ are not dominated by any other gamble. Consequently, the choice set for the maximality criterion is $\{X_1,X_2\}$.
\end{Exp}

\subsection{E-admissibility}
\label{subsec:e-admiss}

As we have seen in the previous section, a gamble $X$ belongs to the choice set according to the maximality criterion if it has a higher expected utility than any other gamble $g'$ for some probability $P$ \emph{that may depend on $g'$}. The \emph{e-admissibility} criterion \cite[page 96]{levi83} strengthens this condition by requiring the existence of a compatible probability $P$ for which $X$ has higher expected utility than any other gamble. Formally, $X$ is in the choice set according to the e-admissibility criterion iff there exists $P$ in $\calP(m)$ such that, for any gamble $X'$, $\esp_{P}(X) \ge \esp_{P}(X')$. This definition results in a choice set that is included in that of the maximality criterion. In \cite{seidenfeld04}, Seidenfeld  compared the maximin and e-admissibility criteria, and argued for the latter in sequential decision problems. We can remark that, in contrast with other decision criteria mentioned until now, e-admissibility defines a choice set directly, without explicitly defining a preference relation. We can, however, construct a preference relation from the choice set, as explained in Section \ref{subsec:defi}. 

The meaning of the e-admissibility criterion seems to be more deeply rooted in the theory of imprecise probability than that of maximality. Moreover, it is much more costly to implement computationally: to determine whether a gamble is e-admissible, we need to solve a linear programming problem \cite{utkin05b,troffaes14}. In the case when uncertainty about the state of nature is described by a mass function $m$, this problem can be formulated using the allocation function (\ref{eq:alloc}). Let $F_1,\ldots,F_q$ denote the focal sets of $m$ (where $q$ can be much smaller than $2^n$), $a_{kj}=a(\omega_k,F_j)$ for all $(k,j)$ such that $\omega_k \in F_j$,   $\a$ a vector containing  all the $a_{kj}$'s, $\p$ the vector of probabilities $\p=(p_1,\ldots,p_s)$, and $\blambda=(\lambda_1,\ldots,\lambda_{i-1},\lambda_{i+1},\ldots,\lambda_n)$ a vector of $n-1$ slack variables. To determine whether gamble $X_i$ is e-admissible, we can solve the following problem:
\[
\min_{\blambda, \a,\p} \;\; \sum_{l\neq i} \lambda_l
\]
subject to:
\begin{eqs}
\sum_{\{k \mid \omega_k \in F_j\}} a_{kj} & = m(F_j), \quad j=1,\ldots,q  \label{eq:LP1}\\
a_{kj} &\ge 0 \quad \forall (k,j): \exists (\omega_k,F_j), \omega_k \in F_j \label{eq:LP2}\\
p_k &= \sum_{j=1}^q a_{kj}, \quad k=1,\ldots,s \label{eq:LP22}\\
\sum_{k=1}^s p_k (u_{ik} - u_{lk})+ \lambda_l& \ge 0, \quad l\neq  i\label{eq:LP3}\\
\lambda_l & \ge 0, \quad l\neq  i. \label{eq:LP4}
\end{eqs}
Eqs. (\ref{eq:LP1}) and (\ref{eq:LP2}) express that the $a_{kj}$'s define an allocation function, and Eq. (\ref{eq:LP22}) expresses that the $p_k$ are compatible probabilities. Eq. (\ref{eq:LP3}) can be written as $\esp_P (X_i) + \lambda_l \ge \esp_P (X_l)$, where $P$ is the probability measure such that $P(\{\omega_k\})=p_k$ for $k=1,\ldots,s$. By minimizing the sum of the $\lambda_l$'s under constraints (\ref{eq:LP1})-(\ref{eq:LP4}), we get the solution $\blambda=\boldsymbol{0}$ iff there exists $P\in \calP(m)$ such that $\esp_P (X_i) \ge \esp_P (X_l)$ for all $l$, i.e., iff gamble $X_i$ is e-admissible. To determine the set of e-admissible gambles, we can start with the choice set of the maximality criterion, and solve the linear program above for each element in that set.

\begin{Exp}
In Example \ref{ex:maximality}, we found that $X_1$ and $X_2$ are in the choice set of the maximality criterion. Let $F_1=\{\omega_1\}$, $F_2=\{\omega_1,\omega_2\}$, $F_3=\{\omega_3\}$ and $F_4=\{\omega_1,\omega_2,\omega_3\}$ be the focal sets of $m$. We recall that $m(F_1)=0.4$, $m(F_2)=0.2$, $m(F_3)=0.1$ and $m(F_4)=0.3$. To find out whether, e.g., $X_1$ is e-admissible, we solve the following linear programming problem:
\[
\min_{\blambda, \a,\p} \;\; \lambda_2+\lambda_3+\lambda_4
\]
subject to:
\begin{align*}
a_{12}+a_{22}&=0.2\\
a_{14}+a_{24}+a_{34}&=0.3\\
p_1 &= 0.4+a_{12}+a_{14}\\
p_2 &= a_{22}+a_{24}\\
p_3 &= 0.1+a_{34}\\
(37-49)p_1+(25-70)p_2+(23-2)p_3 & \ge 0\\
(37-4)p_1+(25-91)p_2+(23-1)p_3 & \ge 0\\
(37-22)p_1+(25-76)p_2+(23-25)p_3 & \ge 0,
\end{align*}
all the variables being positive. We find the solution $\lambda_2=\lambda_3=\lambda_4=0$, $a_{12}=0.2$, $a_{22}=0$, $a_{14}=a_{24}=0$, $a_{34}=0.3$, $p_1=0.2$,  $p_2=0.3$ and $p_3=0.4$. Consequently, gamble $X_1$ is e-admissible. Using the same method, $X_2$ can be shown to be also e-admissible.
\end{Exp}

\section{Shafer's Constructive Decision Theory}
\label{sec:goals}

All the decision criteria reviewed so far rely on the concept of utility. In the axiomatic frameworks developed by von Neumann and Morgenstern in \cite{von_neumann44} as well as  Jaffray in \cite{jaffray89}, utilities are derived from preferences among, respectively, probabilistic and evidential lotteries. In Savage's axiomatic system \cite{savage54}, they are derived from preferences among acts. However, in practice, probabilities (or degrees of belief) and utilities are often elicited from the DM. It is then assumed that probabilities and utilities can be determined independently. Furthermore, the term ``elicitation'' suggests that the DM already has probabilities and utilities in the  back of his mind, and that these values only need to be guessed as accurately as possible. 

In \cite{shafer16d}, Shafer questions these two assumptions. First, he argues, after Savage  \cite[pages 83-84]{savage54}, that when assigning a utility to some consequence corresponding to some way things may turn out, we implicitly ``assess probabilities for how further matters will turn out'' \cite[page 46]{shafer16d}. For instance, to assessing the utility of buying a new car, I need to assess  probabilities for various events such as: the possible withdrawal of my driving license,  various  health problems that could affect my ability to drive, etc. If we assume that predetermined utilities are waiting to be elicited, then it might not matter if utilities are, in fact, expected utilities. However, Shafer questions the existence of preexisting probabilities and utilities, and argues that these values  need to be \emph{constructed}. Probabilities and degrees of belief can be constructed by comparing the problem at hand with a scale of canonical examples such as randomly-coded messages \cite{shafer81} (see Section \ref{subsec:rec_DS}). For utilities, however, it might be difficult or even impossible to ensure that utilities constructed at some level of description are consistent with probabilities and utilities that would be constructed at a more detailed level of description. This is what Shafer calls ``the problem of small worlds''.

\subsection{Formulation of a Decision Problem using Goals} 

Based on the arguments above, Shafer suggested that a constructive decision theory should  be based not on utilities, but on \emph{goals}.  A goal may be defined as a ``consequence''  the DM decides to value and to which he attaches utility irrespective of whatever else happens \cite{shafer16d}. The vocabulary of goals  fits a constructive theory of decision because goals  obviously have to be made. It avoids utility's problem of small worlds as goals constructed at a certain level of description ``by conscious thought and deliberation are the clearest and most definitely structured of all our goals and motives'' \cite{shafer16d}. 

As  explained in Section \ref{subsec:defi}, the standard decision-theoretic framework distinguishes  between a set $\Omega$ of states of nature (or facts about the world that can determine the consequences of our acts), and a set of consequences $\calC$, which specifies how things that the DM cares about may turn out. In contrast, Shafer proposes to use a single frame of discernment $\Theta$,  defined as a  set of collectively exhaustive and mutually exclusive descriptions of how things may turn out. A goal can then be defined as a subset of $\Theta$. Typically, a DM  formulates $n$ goals $A_1,\ldots, A_n$. These goals are \emph{consistent} if their intersection is non-empty, and they are \emph{monotonic} if the subsets are nested, i.e., if for any two goals $A_i$ and $A_j$, we have   either $A_i \subseteq A_j$ or $A_j \subseteq A_i$. One can argue that goals should always be consistent, but two goals initially defined as consistent can become inconsistent after restricting the frame $\Theta$ to a subset $\Theta_0$ as a consequence of acquiring new knowledge. However, monotonic goals $A_1,\ldots,A_n$ can never be made inconsistent after intersection the subsets $A_i$ with some subset $\Theta_0$. 

\subsection{Evaluating Acts}

Assume that performing  act $f$ ensures that things will turn out according to one of the descriptions in some subset $A(f) \subseteq \Theta$. One of the simplest way to evaluate $f$ is to count the number of goals it achieves,
\begin{equation}
u^+(f)=\#\{i \mid A(f) \subseteq A_i\},
\end{equation}
and the number of goals it precludes,
\begin{equation}
u^-(f)=\#\{i \mid A(f) \cap A_i=\emptyset\}.
\end{equation}
We can then assign act $f$ the score
\begin{equation}
U(f)=u^+(f)-u^-(f).
\end{equation}
This method can be extended in two directions. First, we may attach weights $w_1,\ldots,w_n$ to the goals.  The total weight of the goals achieved by  action $f$ is
\begin{equation}
u^+(f)=\sum_{\{i \mid A(f) \subseteq A_i\}} w_i,
\end{equation}
and the total weight of the goals precluded by   $f$ is
\begin{equation}
u^-(f)=\sum_{\{i \mid A(f) \cap A_i=\emptyset\}} w_i.
\end{equation}
As before, the score of $f$ can be defined as $U(f)=u^+(f)-u^-(f)$.

The second important extension is to allow uncertainty in the relation between acts and goal satisfaction. Assume that the effect of act $f$ is represented by a mass function $m_f$ on $\Theta$, with focal sets $F_1,\ldots,F_q$. The expected total weight of goals achieved by  action $f$ is then
\begin{eqs}
\esp\left(u^+(f)\right)&=\sum_{j=1}^q m_f(F_j) \sum_{\{i \mid F_j \subseteq A_i\}} w_i,\\
&=\sum_{i=1}^n w_i \sum_{\{j \mid F_j \subseteq A_i\}} m_f(F_j)\\
&=\sum_{i=1}^n w_i Bel_f(A_i),
\end{eqs}
where $Bel_f$ is the belief function associated to $m_f$. Similarly, the expected total weight of goals precluded by  action $f$ is
\begin{eqs}
\esp\left(u^-(f)\right)&=\sum_{j=1}^q m_f(F_j) \sum_{\{i \mid F_j \cap A_i=\emptyset\}} w_i,\\
&=\sum_{i=1}^n w_i \sum_{\{j \mid F_j \cap A_i=\emptyset\}} m_f(F_j)\\
&=\sum_{i=1}^n w_i Bel_f(\overline{A}_i)\\
&=\sum_{i=1}^n w_i - \sum_{i=1}^n w_i Pl_f(A_i).
\end{eqs}
Dropping the constant term $\sum_{i=1}^n w_i$, the score of $f$ can now be defined as
\begin{equation}
U(f)=\sum_{i=1}^n w_i \left(Bel_f(A_i)+Pl_f(A_i)\right).
\end{equation} 
To see the connection with the MEU principle, we can define  the utility $u(\theta)$ of any  element $\theta$ of $\Theta$ as the total weight of the goals satisfied if $\theta$ holds:
\begin{equation}
u(\theta)=\sum_{\{i \mid \theta\in A_i\}} w_i.
\end{equation}
Let us assume that $m_f$ is Bayesian, and let $p_f(\theta)=m_f(\{\theta\})$ for all $\theta\in \Theta$. Then, the quantities  $u^+(f)$ and $u^-(f)$ become, respectively,
\begin{equation}
u^+(f)=\sum_{\theta\in \Theta} p_f(\theta)u(\theta),
\end{equation}
and
\begin{equation}
u^-(f)=\sum_{i=1}^n w_i - \sum_{\theta\in \Theta} p_f(\theta)u(\theta).
\end{equation}
In that case, the quantity $u^+(f)$ thus becomes the Bayesian expected utility, and $u^-(f)$ is redundant. When the mass functions $m_f$ on $\Theta$ induced by each of the acts $f$  is Bayesian, Shafer's method thus boils down to the MEU criterion, with a suitable definition of the utility function.

\begin{Exp}
\label{ex:Shafer}
As an illustration of the way Shafer's method can be applied in practice, let us consider a classification problem with a set of $K$ classes $\Omega=\{\omega_1,\ldots,\omega_K\}$. Assume that we want to classify an object with unknown class $\bomega\in \Omega$, by selecting a non-empty set $C\subseteq\Omega$ of possible classes. If $\bomega=\omega_k$ and $C=\{\omega_k\}$,  we have a perfectly correct classification. If $\omega_k\in C$ but $\vert C \vert>1$, then the classification is still correct, but imprecise. If $\omega_k\not\in C$, we have an error. The ways things may turn out can be described as follows. On the one hand, the object may actually belong to any of the $K$ classes. On the other hand, we may select a set of cardinality $k$, $k=1,\ldots,K$. If the set has cardinality $K$, then it surely contains the true class; otherwise, we may have a correct classification  or an error. The frame of discernment $\Theta$ can, thus, be defined as follows:
\[
\Theta=\left(\Omega \times \{1,\ldots,K-1\} \times \{{\sf correct},{\sf error}\}\right) \cup \left(\Omega\times \{K\} \times \{{\sf correct}\}\right).
\]
Our general objective is classify the object correctly while being as precise as possible. This objective can be broken down into $K$ monotonic goals $A_1 \subset A_2 \subset \ldots \subset A_K$, where 
\[
A_k=\Omega\times\{1,\ldots,k\}\times\{{\sf correct}\}
\] 
is the goal of selecting a set of at most $k$ elements containing the true class. Goal $A_K$ is to select a set containing the true class, whatever its the size; it can arguably be regarded as the most important, and should be assigned the largest weight. Let $f_C$ denote the act of selecting the non-empty subset $C\subseteq \Omega$, and let $m$ be a mass function on $\Omega$ representing evidence about the class of the object (as provided, for instance, by an evidential classifier such as described in \cite{denoeux00a} or \cite{xu16a}). If $|C|=k$ and $\bomega=\omega_k$, then selecting act $f_C$ will satisfy goals $A_k,\ldots,A_K$ iff $\omega_k \in C$. Consequently, the belief and plausibility of achieving each of the goals $A_k,\ldots,A_K$ when selecting act $f_C$ are, respectively, $Bel(C)$ and $Pl(C)$, and the score of $f_C$ is
\begin{equation}
\label{eq:exShafer}
U(f_C)=\sum_{k=|C|}^K w_k (Bel(C)+Pl(C))=(Bel(C)+Pl(C)) \sum_{k=|C|}^K w_k.
\end{equation}
From (\ref{eq:exShafer}), we can see that $U(f_C)$ is a product of two terms, one of which increases with the size of $C$ due to the monotonicity of mappings $Bel$ and $Pl$, and the other one of which decreases with the size of $C$. 

\begin{table}
\caption{Calculation of the score of acts in Example \ref{ex:Shafer}. \label{tab:Shafer}}
\begin{center}
\begin{tabular}{cccccccc}
\hline
$C$ & $\{\omega_1\}$ & $\{\omega_2\}$ & $\{\omega_1,\omega_2\}$ & $\{\omega_3\}$ &$\{\omega_1,\omega_3\}$ & $\{\omega_2,\omega_3\}$ & $\Omega$\\
\hline
$Bel(C)+Pl(C)$ & 0.8  &  1 & 1.6 & 0.4  &  1 & 1.2 &   2 \\
$\sum_{k=|C|}^K w_k$ & 4 &4& 3& 4& 3& 3 &2\\
$U(f_C)$ & 3.2  &  4 & 4.8 & 1.6 &   3 & 3.6  &  4\\
\hline
\end{tabular}
\end{center}
\end{table}

For instance, assume that $K=3$, $w_1=w_2=1$, $w_3=2$, and let $m$ we the mass function defined as
\[
m(\{\omega_1,\omega_2\})=0.6, \quad m(\{\omega_2,\omega_3\})=0.2, \quad m(\Omega)=0.2.
\]
The calculation of $U(f_C)$ for each non-empty subset $C$ of classes is detailed in Table \ref{tab:Shafer}. We obtain the following preferences among acts:
\[
f_{\{\omega_1,\omega_2\}} \succ f_{\{\omega_2\}} \sim f_\Omega \succ f_{\{\omega_2,\omega_3\}} \succ f_{\{\omega_1\}} \succ f_{\{\omega_1,\omega_3\}} \succ f_{\{\omega_3\}}.
\]
\end{Exp}

\section{Conclusions}
\label{sec:concl}

I have tried in this literature review to provide a broad picture of decision methods applicable to situations where uncertainty about outcomes is formalized in the belief function framework. Interestingly, all methods boil down to MEU when the belief function is Bayesian, but they differ in several important respects in the general case. 

The most important distinction between models is whether they produce a \emph{complete} preference relation or a \emph{partial} one. As shown by Jaffray \cite{jaffray88}, imposing completeness of the preference relation as well as some other requirements  (similar to the Von Neumann and Morgenstern axioms \cite{von_neumann44} in the probabilistic case) leads to  defining the expected utility of an evidential lottery $\mu$ as a weighted sum $\sum_{A\subseteq \calC} \mu(A)U(A)$, where $U(A)$ is the aggregated utility within focal set $A$. The Hurwicz   and  OWA criteria (including the maximin, maximax and pignistic criteria as special cases) are  built on this principle, the  minimax regret being the only notable counterexample.  Smets \cite{smets90e,smets02b,smets05b} was a strong advocate of the  pignistic criterion, which has been widely used in applications.  The main arguments put forward by Smets to support the pignistic criterion are the avoidance of Dutch books\footnote{Snow \cite{snow98} questioned the claim that the transferable belief model, Smets' version of DS theory based on a distinction between credal and pignistic levels, avoids Dutch books. Smets (personal communication) submitted a rebuttal to the \emph{Artificial Intelligence} journal, but this response was never published. It would be interesting to re-examine Snow's arguments and confront them to Smets' views as exposed in various writings.} under forced bets (a case for basing decisions on probability distributions, regardless on the way they are constructed) and the linearity property (\ref{eq:linear_pignistic}), which uniquely determines the pignistic transformation. Smets initially proposed this requirement as an axiom, but it was not generally considered as particularly compelling. In \cite{smets05b}, he derived it from the MEU principle, arguing that the linearity requirement is ``unavoidable provided one accepts expected utility theory''. The argument, however, is complex and would need a critical re-examination. Following a different path, Jaffray \cite{jaffray88} showed  that adding a dominance axiom to complete preorder,  continuity and independence axioms implies that the aggregated utility $U(A)$ within set $A$ should depend only on the utilities of the worst and the best consequences within that set. Accepting this axiom leads us to discarding the pignistic  and OWA criteria. The most general form of decision criterion resulting from Jaffray's axioms is based on locally weighting the minimum and the maximum utility within each focal set using a ``local pessimism index'', generalizing the Hurwicz criterion. We thus have two main methods for building a complete preference relation, supported by different axiomatic arguments: the pignistic and Jaffray's criteria. We can remark that none of these two sets axioms relies on Dempster's rule,  a fundamental building block of DS theory. Work is under way to design a set of axioms supporting a decision criterion with arguments more grounded in DS theory \cite{denoeux19c}.


The other main category of decision models relaxes the assumption of complete preferences and allows incomparability between some acts due to lack of information.  This approach has been particularly studied in the  literature on imprecise probability \cite{troffaes07, huntley14}, because it is in line with the general philosophy of allowing imprecision in an agent's description of uncertainty, and propagating this uncertainty all the way up to the decision level. However, it is also relevant within the DS model. Decision models allowing for incomplete preferences can be further divided into three subcategories: (1)  strong dominance and interval bound dominance  based on lower and upper expected utilities; (2) models based on extensions of the stochastic dominance relation between random variables, and (3) criteria with an imprecise probability flavor (maximality and e-admissibility). More work is needed to evaluate the relevance of maximality and e-admissibility  from the pure DS perspective. One direction might be to consider the set of pignistic probabilities under all refinements, which Wilson \cite{wilson93} showed to be a strict subset of the set of all compatible probabilities. Could we define decision criteria based on that set, and would they be similar to, or different from the criteria derived in the imprecise-probability framework? These are interesting questions that should be addressed in further research.

Finally, Shafer's constructive decision theory, as exposed in a paper written in December 1982 \cite{shafer16a} but only published in 2016 \cite{shafer16d}, departs fundamentally from other approaches and constitutes a category of its own. Shafer questions the practical relevance of the concept of utility, which, he argues, are not pre-existing and waiting to be elicited, but need to be constructed. He proposes to shift the focus from utilities to goals,  formalized as subsets of a frame of discernment comprising both states of nature and ``states of the person'', i.e., consequences of acts. Shafer proposed to score each act by the number of goals they ensure minus the number of goals they preclude. We note that a partial preference relation could also be constructed by considering an act $f_1$ to be at least as desirable as an act $f_2$ if $f_1$ ensures at least as many goals while precluding at most as many goals. As Shafer's decision theory has been overlooked until recently, deeper investigations remain to be carried out to  fully understand its theoretical and practical implications, and to put it in perspective with respect to other approaches.

From this overview of methods of decision-making with belief functions, it is clear that a lot of issues related to decision-making with belief functions remain unsolved and open to investigation. As argued by Bell, Raiffa and Tversky \cite{bell88} (cited in \cite{shafer16d}), decision models can be evaluated \emph{descriptively} by  their empirical validity, \emph{normatively} by their theoretical adequacy, or \emph{prescriptively} by their pragmatic value, i.e., by their ability to help people make better decisions. Very little is known about the value of DS theory as a descriptive model of human reasoning and decision-making under uncertainty, and considerably more work is needed to compare the normative and prescriptive values of the various decision methods reviewed in this paper.

\section*{Ackowledgments}

The authors thanks the two anonymous reviewers for their useful comments. This research was supported by the Labex MS2T, which was funded by the French Government, through the program ``Investments for the future'' by the National Agency for Research (reference ANR-11-IDEX-0004-02).


\section*{References}


\end{document}